%% file: JMLRsample.tex
\documentclass[twoside,11pt]{article}
\usepackage{blindtext}

\usepackage[preprint]{jmlr2e}

\input{math_commands.tex}

\usepackage{hyperref}
\usepackage{enumitem}
\usepackage{url}
\usepackage{multirow}
\usepackage{tabularx}    
\usepackage{makecell}     
\usepackage{tikz}
\usetikzlibrary{arrows.meta, positioning, fit, calc}

\usepackage[ruled,vlined]{algorithm2e}
\usepackage{algpseudocode}
\usepackage{float}
\usepackage{setspace} 
\usepackage{booktabs}

\algrenewcommand\algorithmicrequire{\textbf{Require:}}
\algrenewcommand\algorithmicensure{\textbf{Ensure:}}
\algrenewcommand\algorithmicindent{1.2em} 
\usepackage{lastpage}

\ShortHeadings{A Trainable Centrality Framework for Modern Data}{Vu, Liu and Zhou}
\firstpageno{1}

\begin{document}

\title{A Trainable Centrality Framework for Modern Data}

\author{\name Minh Duc Vu\thanks{Vu and Liu contributed equally to this work.} \email minhduc.vu@u.nus.edu  \\
       \addr Department of Statistics and Data Science\\
       National University of Singapore
       \AND
       \name Mingshuo Liu\footnotemark[1] \email mshliu@ucdavis.edu \\
       \addr Department of Statistics\\
       University of California, Davis
       \AND
       \name Doudou Zhou\thanks{Corresponding author.} \email ddzhou@nus.edu.sg \\
       \addr Department of Statistics and Data Science\\
       National University of Singapore
       }

\editor{My editor}

\maketitle

\begin{abstract}
Measuring how central or typical a data point is underpins robust estimation, ranking, and outlier detection, but classical depth notions become expensive and unstable in high dimensions and are hard to extend beyond Euclidean data. We introduce Fused Unified centrality Score Estimation (FUSE), a neural centrality framework that operates on top of arbitrary representations. FUSE combines a global head, trained from pairwise distance-based comparisons to learn an anchor-free centrality score, with a local head, trained by denoising score matching to approximate a smoothed log-density potential. A single parameter between 0 and 1 interpolates between these calibrated signals, yielding depth-like centrality from different views via one forward pass. Across synthetic distributions, real images, time series, and text data, and standard outlier detection benchmarks, FUSE recovers meaningful classical ordering, reveals multi-scale geometric structures, and attains competitive performance with strong classical baselines while remaining simple and efficient.

\end{abstract}


\section{Introduction}

Quantifying how central or peripheral a data point is relative to a distribution is a core tool in statistics and machine learning. A single centrality score provides a unified way to describe data. It supports robust estimation such as depth-based medians and trimming \citep{tukey1975mathematics}, induces multivariate rankings for visualization and summarization \citep{liu1990notion}, and underlies classical approaches to outlier detection and robust inference \citep{DonGasko1992}. Foundational notions include the spatial ($L_1$) depth and multivariate median \citep{vardi2000multivariate}, spatial quantile depth \citep{serfling2002depth}, and projection depth \citep{zuo2003projection}. Other important centrality measures include Mahalanobis depth \citep{mahalanobis1936}, kernel density centrality \citep{parzen1962estimation}, and potential depth \citep{pokotylo2019classification}. These works have established data depth as a central concept for robust, interpretable multivariate analysis.

Building on these foundations, recent work has extended depth to more general data types. Examples include depths for functional data that handle partially observed trajectories \citep{elias2023integrated}, depths for non-Euclidean or object-valued data \citep{dai2023tukey}, and unified depths for mixed or complex data \citep{blocher2024union}. Such extensions broaden the reach of depth, but also reveal that many computational and inferential challenges persist in modern applications.

Despite their appeal, classical depth notions face several key limitations in modern settings.  
First, many classical depths become impractical computationally or statistically as
the sample size $n$ and dimension $d$ grow. Exact halfspace (Tukey) and projection depths become infeasible beyond moderate $d$ \citep{dyckerhoff2016exact, dyckerhoff2021approximate, fojtik2024exact}, while Mahalanobis depth depends on accurate covariance estimation, which is unreliable when $d$ is comparable to or larger than $n$.  
Second, in high dimensions, pairwise distances concentrate and hubs emerge \citep{beyer1999nearest, radovanovic2010hubs, feldbauer2019comprehensive}, and the data often lie near a low-dimensional manifold \citep{belkin2006manifold, fefferman2016testing, meilua2024manifold}, weakening the meaning of distance-based centrality in raw feature space.  
Third, most classical definitions capture primarily global centrality, describing how a point relates to the overall cloud, while providing limited resolution for local centrality, i.e., how typical a point is within its neighborhood. For example, points near the center of a cluster are intuitively more typical than those at its boundary \citep{rodriguez2014clustering}, yet global centrality may assign them similar scores.  
Finally, traditional notions are tied to hand-crafted formulas and Euclidean geometry, making them difficult to apply directly to modern data types such as images, text, or time series.

A related challenge concerns inference. For many classical depths, evaluating the depth of a new point requires recomputing its relation to the entire reference sample. This coupling prevents fast or real-time inference, since scoring a new point is essentially as expensive as recomputing depths for the full cloud. Only a few special cases, such as Mahalanobis depth, admit closed-form scoring, but these rely on restrictive distributional assumptions and often degrade in high dimensions. In contrast, a learnable model can be trained once and then score new points via a single forward pass, without revisiting the training data, which is an attractive property for large-scale or streaming applications.

These difficulties motivate a more flexible and learnable notion of centrality. In modern machine learning, there is growing interest in distinguishing typical from atypical samples \citep{swayamdipta2020dataset}, often operating in representation spaces learned by self-supervision \citep{chen2020simple} or by manifold-preserving methods such as UMAP \citep{mcinnes2018umap}. In outlier detection, deep one-class models \citep{ruff2018deep} and transformation-based approaches \citep{bergman2020classification} demonstrate the value of adaptive, data-driven centrality. In parallel, score-based generative models \citep{Song2021SDE} and flow-matching techniques \citep{Lipman2023} show that denoising objectives can capture density-related information that scales more gracefully with dimensions than explicit density estimation. Together, these directions suggest that modern centrality should be data-driven, representation-aware, and capable of bridging global and local structures.

However, learning centrality faces a fundamental difficulty. There are no ground-truth labels declaring which points are ``central'' or ``peripheral''. Centrality is inherently relative, and its meaning lies in comparing points, not in assigning them absolute labels. In the absence of labels, existing neural methods typically rely on indirect proxies such as reconstruction error or one-class objectives \citep{ruff2018deep, bergman2020classification}. These proxies capture certain aspects of the data geometry but may not align with depth-like notions, and they often emphasize a global one-class objective by pulling all points toward a single center, at the expense of local density structure.

Our first step is therefore to learn centrality directly from pairwise comparisons. Given two points and a chosen dissimilarity, we declare the one that is closer to the other samples to be more central. Aggregating such
local preferences across many random anchors yields an anchor-free global ordering, in the spirit of ranking and preference learning. This idea underlies our global centrality score. We learn a smooth scalar score on the representation space such that, for any two points, the difference between their scores, passed through a logistic sigmoid, approximates the probability that one would be judged more central than the other when both are compared to random reference points. Conceptually, this learned score acts as a neural analogue of an integrated dissimilarity-based depth.

Global rankings alone are not sufficient, however. Local density structure is crucial for identifying atypical points within clusters or along decision boundaries \citep{foorthuis2021nature, bouman2024unsupervised}. This motivates our second step. Exact nonparametric density estimation is notably difficult in high dimensions, but denoising score matching and related techniques have shown that the gradient of a smoothed log-density is often more stable and learnable than the density itself \citep{hyvarinen2005estimation, vincent2011connection, Song2021SDE}. Intuitively, the score field points towards regions of higher probability and can be viewed as an arrow pointing ``inward'' in the local geometry. When parameterized as the gradient of a scalar potential, this field can be integrated into a function that behaves like a local centrality measure.

We introduce Fused Unified centrality Score Estimation (FUSE), a neural framework for learning centrality across general data types. FUSE operates on fixed representations, and a shared encoder maps these representations to latent features, on top of which we attach two complementary heads. One is a global head, trained via pairwise ranking to approximate an integrated, anchor-free notion of centrality, and the other one is a local head, trained with denoising score matching to approximate a smoothed log-density potential in the latent space.  
Both heads operate in the same representation space and are calibrated onto a common $[0,1]$ scale to make their outputs comparable. A homotopy module $f(x,t)$, indexed by $t\in[0,1]$, then interpolates between the two signals, providing a single interpretable knob that balances global robustness and local sensitivity.

Our contributions are as follows.  
(1) We propose FUSE, a neural centrality framework that combines a ranking-based global signal with a score-based local signal. (2) We introduce a continuous family of centrality scores $f(x,t)$ parameterized by $t\in[0,1]$, providing a clear and interpretable balance between robustness to global structure and sensitivity to local density.  
(3) We show empirically that FUSE recovers classical depth behavior in simple settings while extending naturally to non-Euclidean and high-dimensional data through pretrained encoders.  
(4) Once trained, FUSE scores new points in a single forward pass without revisiting the training samples, enabling fast centrality evaluation suitable for large-scale and streaming scenarios.  
(5) Through extensive experiments, we demonstrate that varying $t$ uncovers structures missed by purely global centrality, yields interpretable orderings in synthetic and real data, and attains competitive outlier detection performance against strong classical baselines.

We use the term ``centrality'' informally. FUSE does not enforce all axioms of classical depth \citep{zuo2000general}, but it retains two key features. Each point is assigned a scalar score, and sorting the sample by their scores induces a one-dimensional ordering that we interpret as going from more central to more outlying points, and any strictly increasing transformation of this score yields the same ranking and level sets. Meanwhile, FUSE extends depth-like ideas to modern, high-dimensional, and non-Euclidean data through representation learning.

The remainder of the paper is organized as follows. Section~2 introduces our neural centrality model and the FUSE model construction. Section~3 presents experiments on synthetic data and real datasets across images, time series, and text, followed by outlier detection tasks. We conclude the paper in Section \ref{sec：conclusion}. Additional experimental results and implementation details are provided in the Supplementary Material. 
A Python implementation of FUSE, together with scripts to reproduce
examples, is available at
\url{https://github.com/vuminhducvmd/FUSE}.

\section{Method}
\label{sec:method}

We aim to learn a scalar centrality score for each point that smoothly interpolates between a
global notion of centrality and a local, density–driven notion. Intuitively, points with high
centrality lie in dense or central regions, whereas points with low centrality lie in sparse or boundary regions and may behave as outliers.

\subsection{Neural centrality model}
Let $\mathcal X$ be a data space, and suppose we observe i.i.d.\ random elements 
$X_1,\dots,X_n \in \mathcal X$ with a common distribution. We write $\mathcal X_n := \{X_1,\dots,X_n\}$ for the observed sample. We start from a fixed representation map
$$
  \psi:\mathcal X \to \mathbb{R}^p,
$$
and define
$$
  Z_i = \psi(X_i) \in \mathbb{R}^p, \quad i=1\ldots,n
$$
as the latent representation of $X_i$, where $p$ is the representation dimension and may vary across architectures.  
The map $\psi$ can be the identity on raw Euclidean data, or a pretrained feature extractor
such as CLIP for images \citep{radford2021learning}, BERT for text
\citep{devlin2019bert}, or SimCLR for visual embeddings \citep{chen2020simple}.  
In all experiments, $\psi$ is kept fixed and is not updated during training, while all
trainable parameters reside in the subsequent shared encoder and the
global and local heads.

We equip the representation space $\mathbb{R}^p$ of $\psi(X)$ with a dissimilarity measure
$
  \delta_Z:\mathbb{R}^p \times \mathbb{R}^p \to [0,\infty),
$
and compute dissimilarities between raw data points via their representations by defining
$
  \delta(X_i,X_j) = \delta_Z\bigl(\psi(X_i),\psi(X_j)\bigr).
$
For Euclidean  data with $\psi$ equal to the identity map, this reduces to applying
$\delta_Z$ directly in the original input space. For non-Euclidean data such as images, text, or
time series, dissimilarities can be evaluated in the representation space through $\psi$.

On top of the fixed latent representation $\psi(x)$ for $x \in \mathcal{X},$ we place a shared encoder
$e_{\theta_e}:\mathbb{R}^p \to \mathbb{R}^q$, followed by two scalar heads with
parameters $\theta_g$ and $\theta_l$.  
For a generic input $x$, we first compute $\psi(x)$ and $e_{\theta_e}\big(\psi(x)\big)$, and then define the global and local heads as
$$
g_\theta(\psi(x)) = h_g\big(e_{\theta_e}\big(\psi(x)\big); \theta_g\big) \in \mathbb{R},
\quad
l_\theta(\psi(x)) = h_l\big(e_{\theta_e}\big(\psi(x)\big); \theta_l\big) \in \mathbb{R}.
$$
Here $\theta = (\theta_e,\theta_g,\theta_l)$ collects all trainable parameters of the
shared encoder and the two heads. For notational simplicity we write $g_\theta$ and
$l_\theta$ to denote the resulting global and local scoring functions.

The global head $g_\theta$ is trained to produce an anchor–free ranking that aligns with
pairwise centrality comparisons, capturing coarse, cross–cluster structure. The
local head $l_\theta$ is trained via denoising score matching (DSM)
\citep{hyvarinen2005estimation,vincent2011connection} so that its gradient approximates the score
of a Gaussian–smoothed version of the data distribution in representation space, capturing
density–driven local typicality. In Section~\ref{sec:homodepth} we introduce a homotopy module
$f(x,t)$, $t\in[0,1]$, that interpolates between these two calibrated signals to yield a
continuous family of centrality scores controlled by a single parameter $t$.

\subsubsection{Global head: anchor–marginal ranking}
\label{sec:global}

The global head learns an anchor–free centrality score $g_\theta(\psi(x))$ such that points closer to
the majority of the data receive larger scores. Our starting point is that centrality is inherently
comparative. Given two candidates $x_1,x_2\in\mathcal X$ and a reference point $x_0\in\mathcal X$,
we can say which one is more central relative to $x_0$ based on distances under $\delta$, even
though we do not observe absolute labels.

For an anchor $x_0$ and candidates $x_1,x_2$, we define the binary relation
$$
x_1 \prec_{x_0} x_2
\quad\Longleftrightarrow\quad
\delta(x_1,x_0)
<
\delta(x_2,x_0),
$$
meaning that, relative to $x_0$, $x_1$ is more central than $x_2$ if it is closer in the
representation space. This comparison rule is deterministic given $(x_0,x_1,x_2)$. To remove the dependence on any particular anchor, we marginalize over the random anchor $X_0$.
For a fixed pair $(x_1,x_2)\in\mathcal X\times\mathcal X$, we define the marginal preference
probability
$$
\pi(x_1,x_2)
:=
\Pr\big[x_1 \prec_{X_0} x_2\big],
$$
which is the probability that $x_1$ is judged more central than $x_2$ when compared to a random
reference point $X_0$.

In practice, we approximate this probability using a finite anchor set
$S_0 \subset \mathcal X_n$. At each epoch we partition the sample into an anchor set $S_0$ and two
candidate sets $S_1,S_2$, and draw a moderate collection of candidate pairs
$\mathcal S_{12} \subset S_1\times S_2$. For each $(x_1,x_2)\in \mathcal S_{12}$ we then estimate
$$
\hat\pi(x_1,x_2)
=
\frac{1}{|\mathcal A_{12}|}
\sum_{x_0\in \mathcal A_{12}}
\mathbf 1\{x_1 \prec_{x_0} x_2\},
$$
where $\mathcal A_{12}\subset S_0$ is a subset of anchors and $|\cdot|$ denotes
cardinality. This partitioned scheme greatly reduces the cost of computing preferences while
still providing a rich set of comparisons. Concrete sampling choices and further variants are
described in Section~\ref{sec:training} and Supplement~\ref{sec:diff,sample}.

To map empirical preference probabilities to a scalar score, we adopt the Bradley--Terry (BT)
model for paired comparisons \citep{bradley1952rank}. The BT model assumes that each item $x$ has
a one–dimensional latent score $g_\theta(\psi(x))$ and that the probability of $x_1$ being preferred to
$x_2$ is a logistic function of the score difference
$$
\Pr\big[x_1 \text{ preferred to } x_2\big]
\approx
\sigma\big(g_\theta(\psi (x_1))-g_\theta(\psi (x_2))\big),
$$
where $\sigma(u) = (1 + e^{-u})^{-1}$ is the logistic sigmoid. In our setting, “preferred” means “judged more
central than” when compared via random anchors. Thus $g_\theta(\psi(x))$ can be interpreted as a
global centrality score, and larger $g_\theta(\psi(x))$ means that $x$ tends to win more comparisons
against other points.

We train $g_\theta$ by minimizing the cross–entropy between the empirical preference
$\hat\pi(x_1,x_2)$ and the BT model prediction,
\begin{equation}
\label{eqn:globalloss}
\begin{aligned}
\mathcal L_{\mathrm{global}}(\theta)
=
\mathbb E_{(X_1,X_2)}\Big[
 & -\hat\pi(X_1,X_2)\,
   \log\sigma\big(g_\theta(\psi(X_1))-g_\theta(\psi(X_2))\big) \\
 & -\big(1-\hat\pi(X_1,X_2)\big)\,
   \log\big(1-\sigma(g_\theta(\psi(X_1))-g_\theta(\psi(X_2)))\big)
\Big],
\end{aligned}
\end{equation}
and the expectation is approximated by the empirical average over
$\mathcal S_{12}$. For each pair $(x_1,x_2) \in \mathcal S_{12}$, $\hat\pi(x_1,x_2)$ is the fraction of anchors that consider
$x_1$ more central than $x_2$, while
$\sigma\big(g_\theta(\psi(x_1))-g_\theta(\psi(x_2))\big)$ is the model's predicted probability. The loss
encourages $g_\theta(\cdot)$ to preserve these integrated orderings of centrality.

Anchors are used only to construct supervision during training. After marginalization, the
resulting score $g_\theta(\psi(x))$ depends solely on $x$ and can be evaluated independently at
inference time via a single forward pass, yielding an anchor–free global centrality score whose
pairwise differences approximate integrated depth–like preferences.

\subsubsection{Local head: denoising score matching}
\label{sec:local}

The global head describes how a point behaves with respect to the dataset as a whole and mainly
captures coarse, cross–cluster structure. In clustered data, we often observe that points from
the same cluster receive similar global scores even when one lies near the cluster center and
another near the boundary, so the global score is less sensitive to fine within–cluster
differences. This role is similar to that of classical global depth functions, which provide a
center–outward ordering for the full distribution but are not designed to capture local,
cluster–specific structure \citep{agostinelli2011local,francisci2023analytical}. For tasks such
as outlier detection within clusters, we therefore also need a notion of local centrality that is
sensitive to nearby density \citep{bouman2024unsupervised}, in line with local depth constructions and local–density based
outlier methods such as local outlier factor \citep{breunig2000lof} and cluster-based local
outlier detection \citep{he2003discovering}.

A natural way to formalize local centrality is through the density of the representation. Recall $Z=\psi(X)$ and let $p_Z$ be its density function. Points in regions where $p_Z$ is large are more central locally. Directly
estimating $p_Z$ is difficult in high dimensions, but score–based methods circumvent this by
focusing on the gradient of a smoothed log–density, which is typically more stable and learnable
than the density itself \citep{hyvarinen2005estimation,vincent2011connection,Song2021SDE}. The
resulting score field points towards regions of higher probability and therefore encodes how
typical a point is within its local neighborhood.

For a noise scale $\eta>0$, we define the Gaussian–smoothed density
$$
p_{Z,\eta}(z) := (p_Z * \varphi_\eta)(z),
\quad
s_\eta(z) := \nabla_z \log p_{Z,\eta}(z),
$$
where $\varphi_\eta$ is the density of $\mathcal N(0,\eta^2 I_p)$ and $*$ denotes
convolution. We parameterize a scalar potential $l_\theta:\mathbb R^p\to\mathbb R$ and define the
model score
$$
s_\theta(z) := \nabla_z l_\theta(z),
$$
which is meant to approximate $s_\eta(z)$.

Denoising score matching (DSM) \citep{vincent2011connection} provides a principled and scalable
way to train this model.  Consider the joint density of a pair $(z,\tilde z)$,
$$
q_\eta(z,\tilde z) = p_Z(z)\,\varphi_\eta(\tilde z - z).
$$
Let $(Z,\tilde Z)\sim q_\eta$ and write $q_{\eta}(\tilde z|z)$ as the corresponding conditional density.  Following \citet[Eq.~(9)]{vincent2011connection}, the DSM
objective for a score field $s_\theta(\tilde z)$ can be written as
$$
\mathcal L_{\mathrm{local}}(\theta)
= \frac{1}{2} \mathbb E_{(Z, \tilde Z)}\!\Big[
\big\|
s_\theta(\tilde Z)
-
\partial_{\tilde z}\log q_\eta(\tilde Z \mid Z)
\big\|_2^2\Big].
$$
For Gaussian corruption, \citet[Eq.~(10)]{vincent2011connection} shows that
$\partial_{\tilde z}\log q_\eta(\tilde z \mid z) = \eta^{-2}(z-\tilde z)$.  Writing
$\varepsilon := \tilde Z - Z$, then we have
$\partial_{\tilde z}\log q_\eta(\tilde Z \mid Z) = -\eta^{-2}\varepsilon$, and we obtain the equivalent form
\begin{align}
\mathcal L_{\mathrm{local}}(\theta)
&=
\frac{1}{2} \mathbb E_{(Z,\varepsilon)} \Big[
\big\|
s_\theta(Z+\varepsilon) + \eta^{-2}\varepsilon
\big\|_2^2 \Big],
\label{eqn:local,loss}
\end{align}
where $Z$ and $\varepsilon \sim \mathcal N(0, \eta^2 I_p)$ are independent. In implementation, the expectation in \eqref{eqn:local,loss} is approximated by averaging over the
training representations $\{Z_i\}_{i=1}^n$ and a small number of Gaussian noise samples per point, with details in Section~\ref{sec:training}. Minimizing $\mathcal L_{\mathrm{local}}$ over a sufficiently
rich class encourages $s_\theta$ to approximate the smoothed score
$s_\eta$ \citep{hyvarinen2005estimation,vincent2011connection}, so that $l_\theta$ behaves like
a smoothed log-density up to an additive constant.  We therefore interpret $l_\theta(z)$ as a
local centrality function. Larger values indicate that $z$ lies in a region where the smoothed
density $p_{Z,\eta}$ is high, while smaller values correspond to boundary or low-density
regions.

On the sample $\{Z_i\}_{i=1}^n$, we define the empirical local centrality score as
$$
f_l(X_i) := \mathrm{Normalize}\big(l_\theta(Z_i)\big),
$$
where the function $\mathrm{Normalize}(\cdot)$ maps values to $[0,1]$, and details can be seen in Section \ref{sec:homodepth}. This construction can be viewed as a learned, high–dimensional
analogue of Gaussian kernel density estimation \citep{parzen1962estimation,
silverman2018density}. Instead of explicitly summing kernels around each point, DSM learns a
parametric potential $l_\theta$ whose gradient matches the smoothed score field $s_\eta$. The
resulting model is teacher–free, efficient at inference, and complements the global head by
capturing fine–grained, density–driven typicality.

\subsection{FUSE: homotopy centrality interpolation}
\label{sec:homodepth}

The global and local heads provide two complementary centrality signals, but their raw outputs
live on different scales and are not directly comparable. To put them on a common scale and
expose a single interpretable control over their relative influence, we first calibrate both
outputs to the interval $[0,1]$ and then combine them via a homotopy.

For the global head, $g_\theta(\psi(x))$ plays the role of a BT logit. Since the training loss is
logistic, applying a sigmoid transformation
$$
f_g(X) = \sigma\big(g_\theta(\psi(X))\big)
$$
is natural, since it maps $g_\theta(\psi(X))$ to $[0,1]$ and yields a probability–like global centrality
score. Any strictly increasing transformation would induce the same ranking, but $\sigma(\cdot)$ is numerically stable and interpretable.

For the local head, $l_\theta(\psi(X))$ serves as a surrogate smoothed log–density up to an
additive constant. Because such log–densities can be shifted arbitrarily, we normalize
$l_\theta(X)$ empirically. Let $q_{0.01}$ and $q_{0.99}$ denote the empirical 1st and 99th
percentiles of $\{l_\theta(\psi(X_i))\}_{i=1}^n$ on the training set. We define the local centrality
score
$$
f_l(X)
=
\min\Bigl\{1,\ \max\Bigl\{0,\,
\frac{l_\theta(\psi(X)) - q_{0.01}}{\,q_{0.99}-q_{0.01}\,}
\Bigr\}\Bigr\},
$$
which linearly rescales $l_\theta(\psi(X))$ between these quantiles and clips extreme values.
This normalization makes the local scores comparable to the global scores and robust to
outliers. During inference, $q_{0.01}$ and $q_{0.99}$ are kept fixed.

We then form the homotopy centrality score by linear interpolation,
$$
f(X,t) = (1-t)\,f_g(X) + t\,f_l(X), \qquad t\in[0,1].
$$
The parameter $t$ controls the balance between the two signals: $t=0$ yields a purely global
centrality score, $t=1$ yields a purely local one, and intermediate values provide mixtures of
global and local behavior. We adopt linear interpolation for simplicity and interpretability,
while other schedules are also possible.

\subsection{Training}
\label{sec:training}

\begin{figure}[!htbp]
    \centering
    \includegraphics[width=0.75\linewidth]{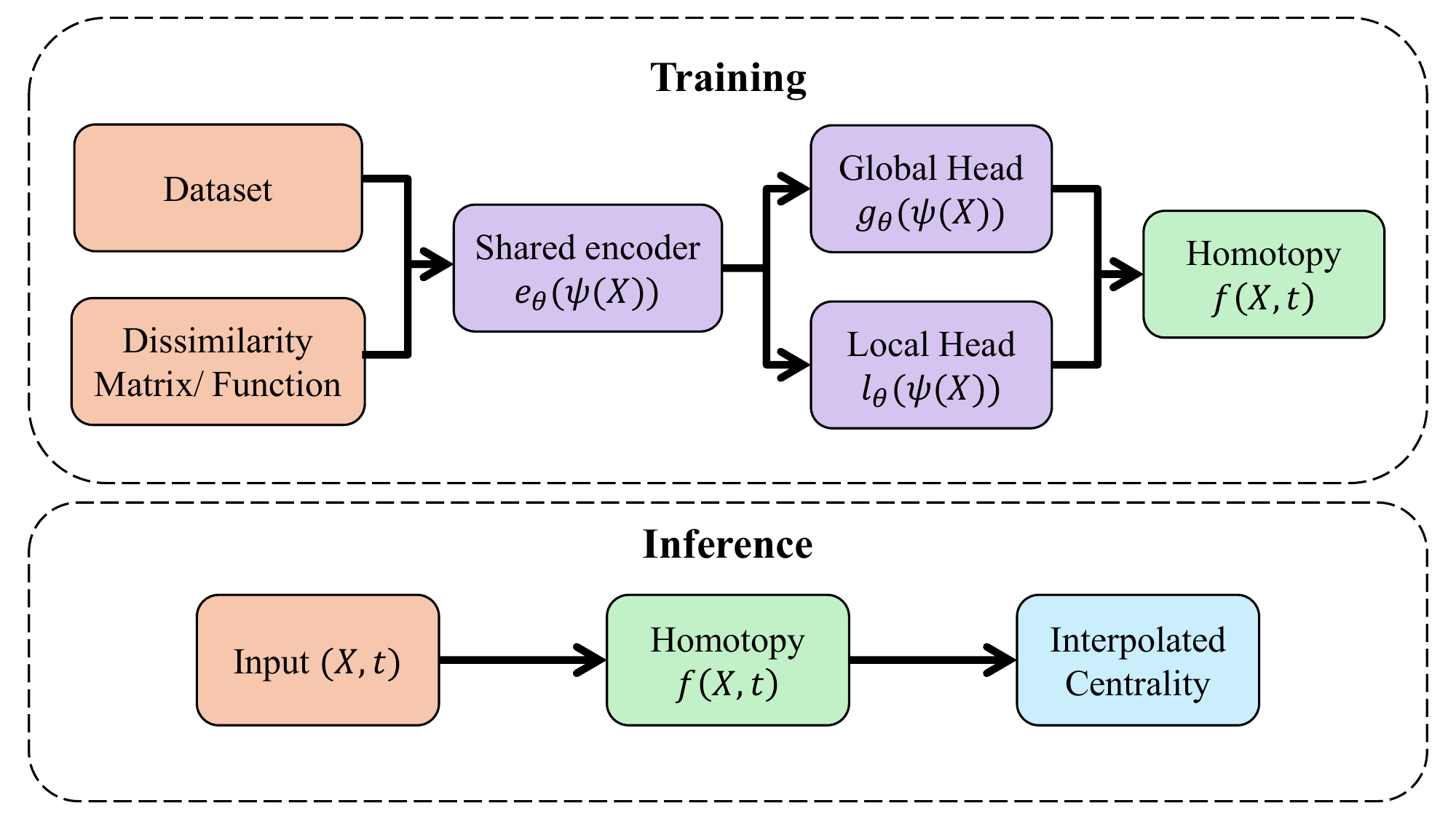}
    \caption{Training (top): mapped data $\psi(X)$ pass through a shared encoder and branch into global
    $g_\theta$ and local $l_\theta$ heads. Inference (bottom): given $(X,t)$, the homotopy
    outputs a single centrality score $f(X,t)$. The dissimilarity $\delta$ can be defined on raw
    data or pretrained embeddings.}
    \label{fig:diagram}
\end{figure}

The overall training alternates between the global pairwise–ranking loss
\eqref{eqn:globalloss} and the local DSM loss \eqref{eqn:local,loss}, allowing the model to
capture both coarse global structure and fine–grained local density. The workflow is summarized
in Figure~\ref{fig:diagram}, and detailed pseudocode is given in
Algorithm~\ref{alg:fit}.

\begin{algorithm}[htbp]
\caption{\textsc{Training}: Joint training of global and local centrality heads}
\label{alg:fit}
\DontPrintSemicolon
\SetKwInOut{Input}{Input}
\SetKwInOut{Output}{Output}

\Input{Sample $\mathcal{X}_n = \{X_i\}_{i=1}^n$; representation map $\psi$; dissimilarity
$\delta_Z$; number of epochs $T$; DSM noise scale $\eta>0$; number of
DSM resamples per epoch $R\in\mathbb{N}$; optimizer $\mathcal{O}$.}
\Output{Trained global and local heads $g_\theta$ and $l_\theta$.}

  \tcp{Precompute representations and induced dissimilarity}
  Define $\delta(X_i,X_j) := \delta_Z(\psi(X_i), \psi(X_j))$ for use below.\;

\For{$\text{epoch} \gets 1$ \KwTo $T$}{

  Randomly split $\mathcal{X}_n$ into three disjoint subsets $S_0,S_1,S_2$.\;

  \tcp{Global head: anchor-marginal ranking}
  Sample a set of candidate pairs $\mathcal{S}_{12} \subset S_1\times S_2$ of size $\lfloor n/3\rfloor$.\;
  \ForEach{$(X_i,X_j) \in \mathcal{S}_{12}$}{
    Sample a small anchor set $\mathcal{A}_{ij}\subset S_0$.\;
    Set $\hat\pi(X_i,X_j)$ to the fraction of anchors $X_0\in\mathcal{A}_{ij}$ such that
    $\delta(X_i,X_0) < \delta(X_j,X_0)$.\;
  }
  Compute the global loss $\mathcal{L}_{\mathrm{global}}(\theta)$
  using~\eqref{eqn:globalloss} with pairs $(X_i,X_j)$ and labels $\hat\pi(X_i,X_j)$, and update
  $\theta$ with optimizer $\mathcal{O}$.\;

  \If{$R>0$}{
    \tcp{Local head: denoising score matching}
    \For{$r\gets1$ \KwTo $R$}{
      \For{$i\gets1$ \KwTo $n$}{
        Set $Z_i \gets \psi(X_i)$, draw $\varepsilon_i \sim \mathcal{N}(0,\eta^2 I_p)$, and form
        noisy input $\tilde Z_i \gets Z_i + \varepsilon_i$.\;
      }
      Compute scores $s_\theta(\tilde Z_i) = \nabla_{z} l_\theta(z)\big|_{z=\tilde Z_i}$ and the
      empirical DSM loss $\mathcal{L}_{\mathrm{local}}(\theta)$
      from~\eqref{eqn:local,loss}, then update $\theta$ with optimizer $\mathcal{O}$.\;
    }
  }
}

Compute calibration quantiles $q_{0.01},q_{0.99}$ from
$\{l_\theta(\psi(X_i))\}_{i=1}^n$ on the training set and fix them for inference.\;

\end{algorithm}

\begin{algorithm}[htbp]
\caption{\textsc{Inference}: Interpolated homotopy centrality score}
\label{alg:infer}
\DontPrintSemicolon
\SetKwInOut{Input}{Input}
\SetKwInOut{Output}{Output}

\Input{New point $X_{\mathrm{new}}$; homotopy parameter $t \in [0,1]$; representation map $\psi$; heads $g_\theta,l_\theta$; calibration
quantiles $q_{0.01},q_{0.99}$.}
\Output{Interpolated centrality $f(X_{\mathrm{new}}, t)$.}

Compute global score
$f_g(X_{\mathrm{new}}) \gets \sigma\big(g_\theta(\psi(X_{\mathrm{new}}))\big).$

Compute local score
$$
f_l(X_{\mathrm{new}}) \gets
\min\Bigl\{1,\ \max\Bigl\{0,\,
\frac{l_\theta(\psi(X_{\mathrm{new}})) - q_{0.01}}{\,q_{0.99}-q_{0.01}\,}
\Bigr\}\Bigr\}.
$$

Return interpolated centrality
$$
f(X_{\mathrm{new}}, t) \gets (1-t) f_g(X_{\mathrm{new}}) + t f_l(X_{\mathrm{new}}).
$$\;

\end{algorithm}

During training, the global ranking loss \eqref{eqn:globalloss} is implemented using the
partitioned sampling scheme introduced in Section~\ref{sec:global}. At each epoch, we construct three disjoint subsets $(S_0,S_1,S_2)$ of $\mathcal X_n$ to
approximate the expectations in \eqref{eqn:globalloss}. Specifically, we randomly split the
sample into an anchor set $S_0$ and two candidate sets $S_1$ and $S_2$ of (approximately) equal size. From $S_1$ and $S_2$, we
sample $\lfloor n/3\rfloor$ pairs $(x_1,x_2)$ to compose $\mathcal{S}_{12}$, and for each pair, sample a small number of
anchors (64 in our experiments) from $S_0$ to estimate the empirical preference $\hat\pi(x_1,x_2)$. This partitioned scheme provides a good trade–off between computational cost and performance,
yielding comparable accuracy with substantially lower time complexity. Alternative sampling strategies and comparisons are discussed in Supplement~\ref{sec:diff,sample}.

Jointly, the local potential $l_\theta$ is optimized using DSM. At each epoch, for each
representation $Z_i$ we generate a fixed number 8 of Gaussian perturbations to approximate the
expectation in \eqref{eqn:local,loss} and update the model using the resulting noisy batch. In
our synthetic and first four real data experiments we fix the noise scale at $\eta=1$, while in the
outlier–detection benchmarks we treat $\eta$ as a tuning parameter. The numbers
$\lfloor n/3\rfloor$, 64, and the number of DSM resamples per epoch are chosen for illustration. When time and computational resources permit, increasing these values yields more accurate Monte
Carlo estimates of the underlying expectations.

After training, we fix the network parameters $\theta$ and the calibration constants
$q_{0.01},q_{0.99}$ estimated on the training set, and we do not update them during inference.
Given any new point $X_{\mathrm{new}}$, FUSE computes its global score $f_g(X_{\mathrm{new}})$ and
local score $f_l(X_{\mathrm{new}})$ via a single forward pass followed by the fixed
normalization in Section~\ref{sec:homodepth}, and then evaluates $f(X_{\mathrm{new}},t)$ for any
$t\in[0,1]$. The detailed pseudocode is given in Algorithm \ref{alg:infer}. In contrast to many classical depth functions that must recompute a point's
relation to the entire reference sample, FUSE performs fast inference. Once training is complete,
scoring new points requires no access to the training data and no further optimization, making
the method well suited to large–scale or streaming applications.

Finally, all experiments use a small shared encoder feeding the two scalar heads $g_\theta$ and
$l_\theta$, and the architecture choices for each dataset are summarized in
Supplement~\ref{app:net-arch}.

\section{Experiments}
\label{sec:experiments}

We evaluate FUSE on synthetic and real-world datasets to assess three key questions:  
(i) whether the learned scores reproduce qualitative behaviors expected from classical centrality or depth notions;  
(ii) whether the homotopy parameter $t\in[0,1]$ provides an interpretable interpolation between global and local structure; and  
(iii) whether FUSE achieves competitive performance on outlier detection benchmarks. Notably, $t{=}0$ corresponds to the global regime (Global) and $t{=}1$ corresponds to the local regime (Local).

We begin with controlled synthetic experiments to examine qualitative behavior and monotonicity.  
We then visualize FUSE on three representative modalities, including images, time series, and text, to illustrate how $t$ reveals multi-scale structures.  
Finally, we evaluate FUSE and its global and local components on two outlier detection benchmarks.

\subsection{Synthetic data}
\label{sec:exp-synth}

Our synthetic experiments test whether FUSE recovers the expected center–outward behavior on standard distributions and whether the homotopy parameter yields a smooth transition from global to local structures.

For illustration, each setting contains $n = 5000$ i.i.d.\ samples in two dimensions. 
In the Gaussian setting we draw
$X \sim \mathcal{N}(0,\Sigma)$, where the covariance matrix
$\Sigma \in \mathbb{R}^{2\times 2}$ is a random symmetric
positive-definite matrix generated by
\texttt{sklearn.datasets.make\_spd\_matrix}. For the Student-$t$ case we sample each coordinate independently from
a centered $t_{10}$ distribution, and for the Uniform case we sample
each coordinate independently from $\mathrm{Unif}(-2,2)$. For such unimodal families, we expect: (i) points near the center to receive larger centrality values;  
(ii) scores decrease monotonically as we move outward; and  
(iii) high correlation between centrality and Euclidean distance to the population center.  
We use these settings to assess how the homotopy scores $f(x,t)$ capture classical depth-like behavior. We next consider a multimodal setting using a Gaussian mixture with four components. Global centrality should emphasize coarse, cross-cluster structure, assigning similar values within each component, and  
local behavior should highlight cluster boundaries and mode-specific density.  
By sweeping $t$ from $0$ to $1$, we examine whether FUSE progressively resolves individual clusters while preserving global structure. Euclidean distances are used to construct the FUSE model.

\begin{figure}[!htbp]
    \centering
    \includegraphics[width=1.0\linewidth]{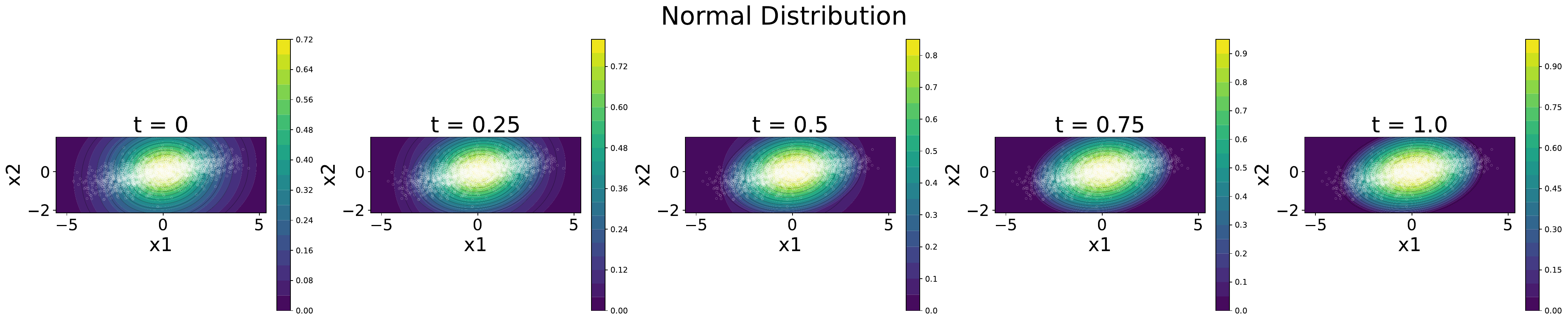}\\[2pt]
    \includegraphics[width=1.0\linewidth]{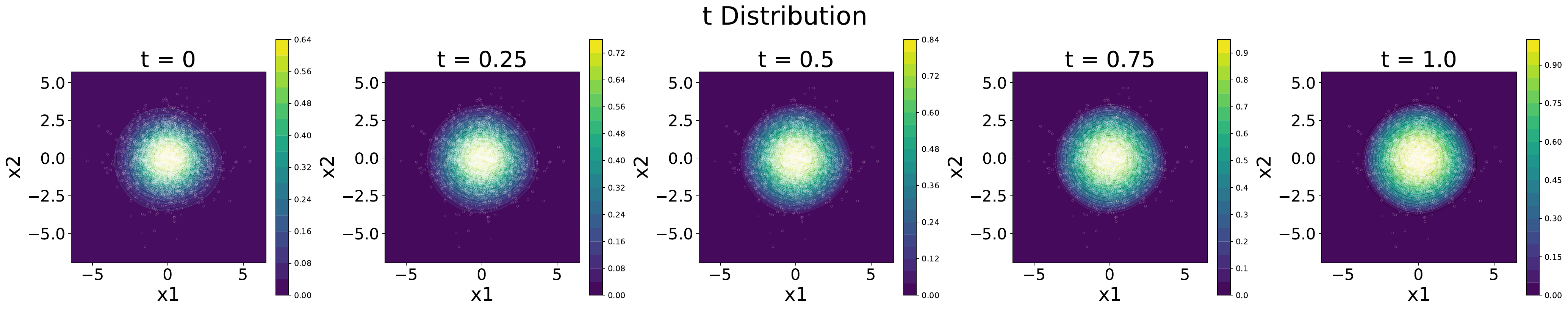}\\[2pt]
    \includegraphics[width=1.0\linewidth]{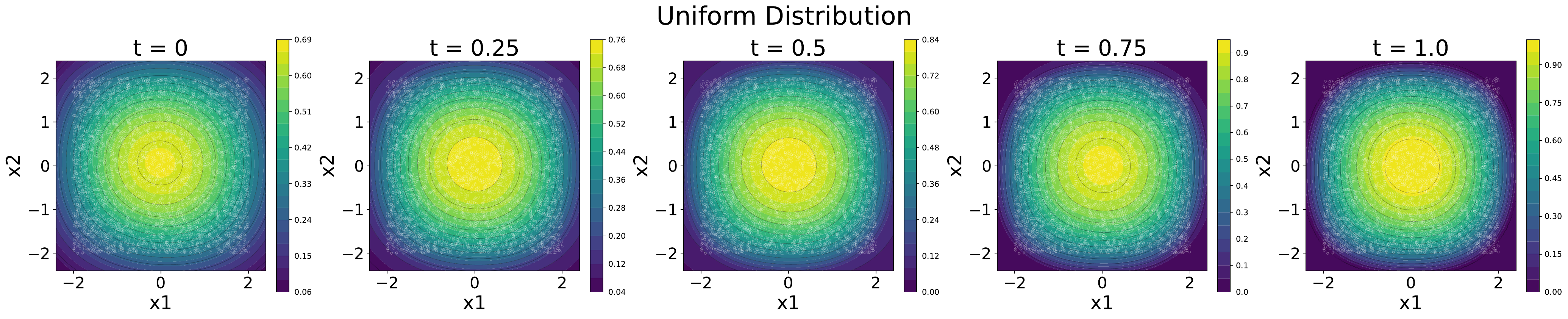}\\[2pt]
     \includegraphics[width=1.0\linewidth]{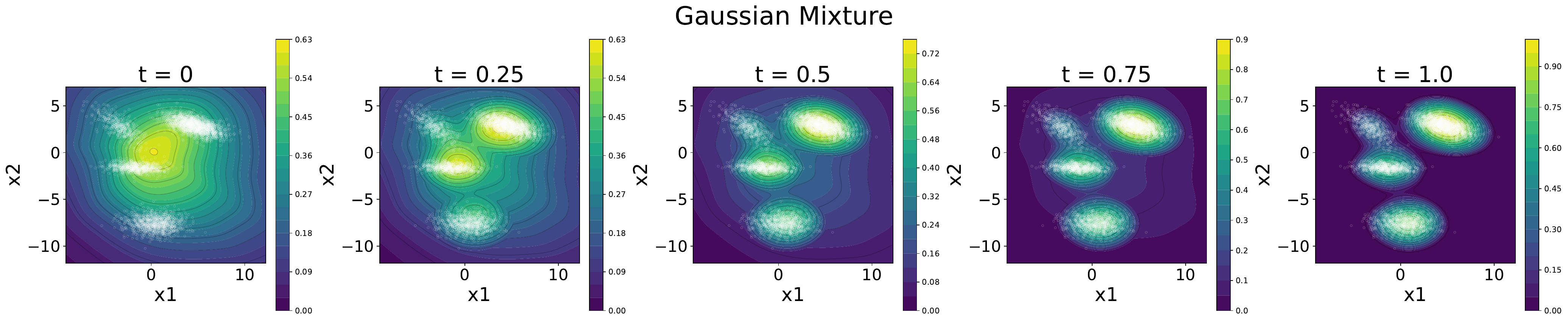}
    \caption{From top to bottom: Homotopy centrality contours on Normal, Student-$t$, Uniform, Gaussian mixture. $t \in \{0, 0.25, 0.5, 0.75, 1\}$.}
    \label{fig:unimodol}
\end{figure}

Figure~\ref{fig:unimodol} shows the learned homotopy centrality surfaces for $t\in\{0,0.25,0.5,0.75,1\}$.  
For Gaussian and Student-$t$ distributions, centrality peaks at the mean and decays smoothly outward, matching elliptical level sets of the density.  
For the Uniform distribution, although the density is constant, FUSE still learns a meaningful notion of center versus boundary. For the Gaussian mixture, $t$ creates a smooth path from a global basin spanning all components to mode-specific local structure. These behaviors align closely with the intuition behind classical depth.

We compare Global and Local against classical centrality measures, including
Kernel Density Estimation (KDE; \citealt{parzen1962estimation}),
Mahalanobis depth (MAH; \citealt{mahalanobis1936}), potential depth (POT; \citealt{pokotylo2019classification}), projection depth (PRO; \citealt{zuo2000general}), 
spatial depth (SPA; \citealt{vardi2000multivariate}) and Tukey depth (Tukey; \citealt{tukey1975mathematics})
All methods are implemented following their reference descriptions and parameters
as described in Supplement~\ref{app:otherdepth}. Since the scales of different
methods' outputs are not directly comparable, we focus on rank-based metrics
and compute Spearman’s and Kendall’s rank correlations
\citep{spearman1904proof,kendall1938new} between the learned centrality
scores and the Euclidean distance to the mean.

\begin{table}[!htbp]
\centering
\caption{Rank correlations between centrality and Euclidean distance to the mean.}
\renewcommand{\arraystretch}{1.15}
\setlength{\tabcolsep}{4.5pt}
\begin{tabular*}{\linewidth}{@{\extracolsep{\fill}}llcccccccc}
\toprule
\textbf{Data} & \textbf{Metric} &
\textbf{Global} & \textbf{Local} & \textbf{KDE} &
\textbf{MAH} & \textbf{POT} & \textbf{PRO} & \textbf{SPA} & \textbf{Tukey} \\
\midrule
\multirow{2}{*}{Normal}
 & Spearman & 0.996 & 0.982 & 0.981 & 0.781 & 0.999 & 0.776 & 0.993 & 0.782 \\
 & Kendall  & 0.951 & 0.894 & 0.894 & 0.603 & 0.981 & 0.598 & 0.940 & 0.605 \\
\midrule
\multirow{2}{*}{Student-$t$}
 & Spearman & 0.998 & 0.999 & 1.000 & 1.000 & 1.000 & 0.999 & 1.000 & 1.000 \\
 & Kendall  & 0.964 & 0.980 & 0.992 & 0.991 & 0.997 & 0.977 & 0.990 & 0.982 \\
\midrule
\multirow{2}{*}{Uniform}
 & Spearman & 0.998 & 0.998 & 0.998 & 1.000 & 1.000 & 0.988 & 0.999 & 0.989 \\
 & Kendall  & 0.966 & 0.961 & 0.962 & 0.984 & 0.966 & 0.915 & 0.970 & 0.926 \\
\bottomrule
\end{tabular*}
\label{tab:syn}
\end{table}

Table~\ref{tab:syn} reports these correlations. Both Global and Local achieve
the highest or near-highest values on the unimodal settings, with Spearman and
Kendall correlations typically above $0.95$, demonstrating a strong monotone
alignment with the ground-truth structure. The mixture example further
highlights FUSE’s flexibility. By varying $t$, FUSE smoothly transitions from a global to a local centrality view.

\begin{figure}[!ht]
    \centering
    \includegraphics[width=0.48\linewidth]{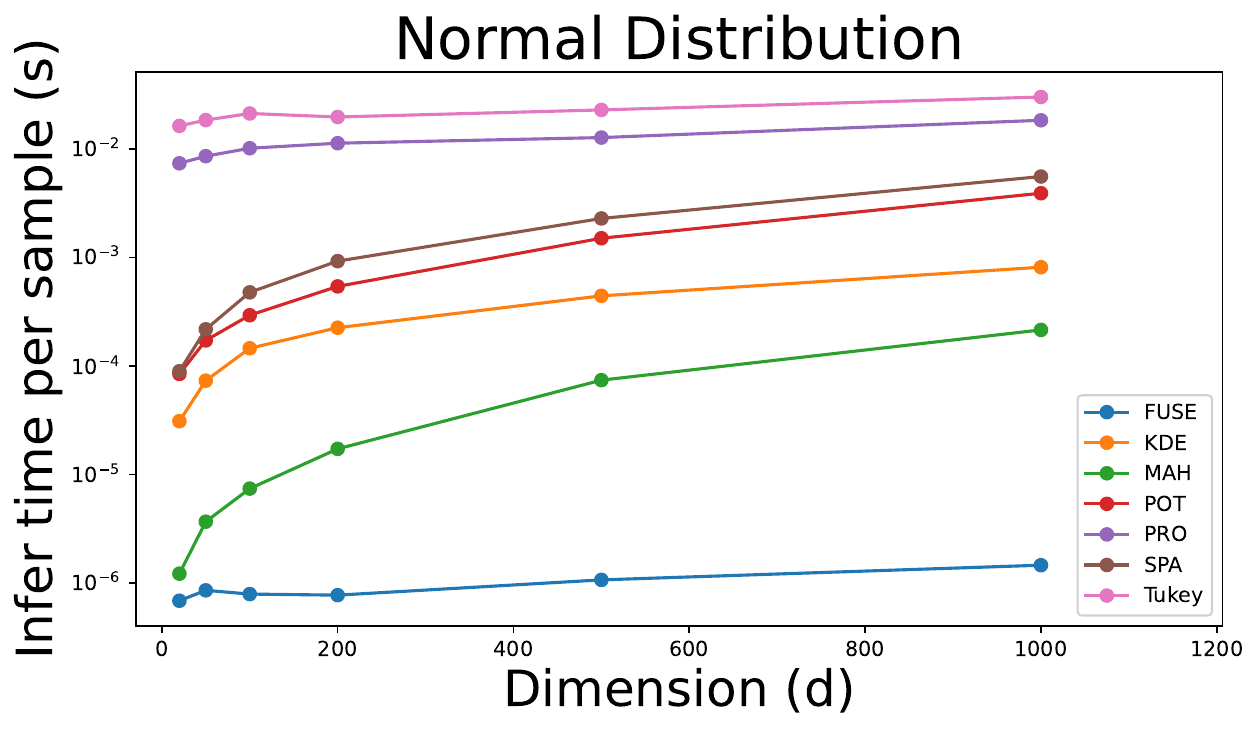}\hfill
    \includegraphics[width=0.48\linewidth]{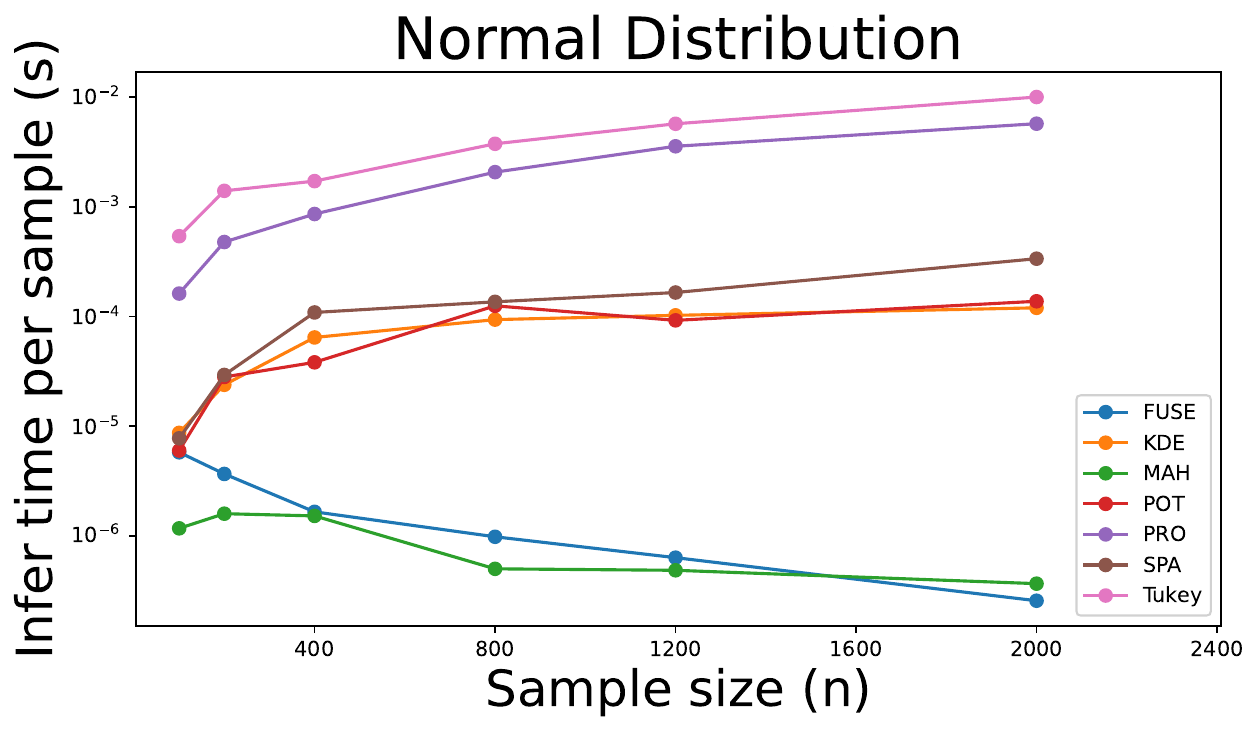}\\[4pt]
    \includegraphics[width=0.48\linewidth]{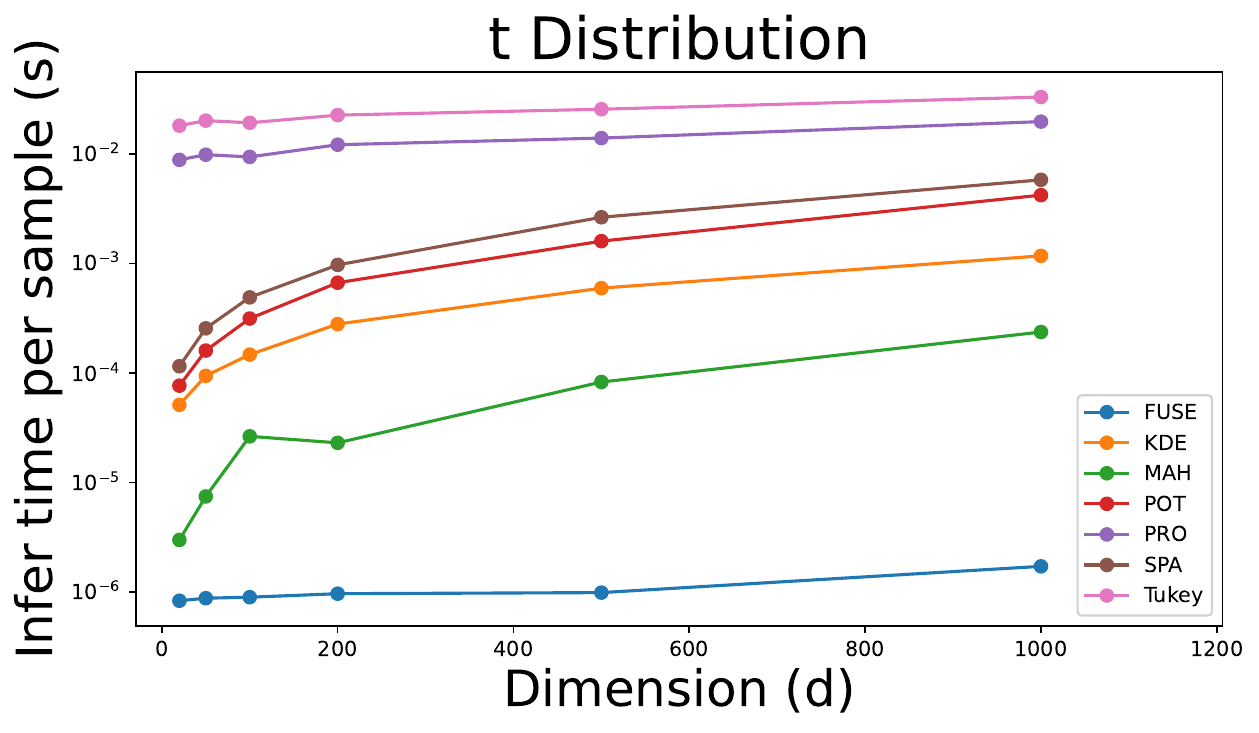}\hfill
    \includegraphics[width=0.48\linewidth]{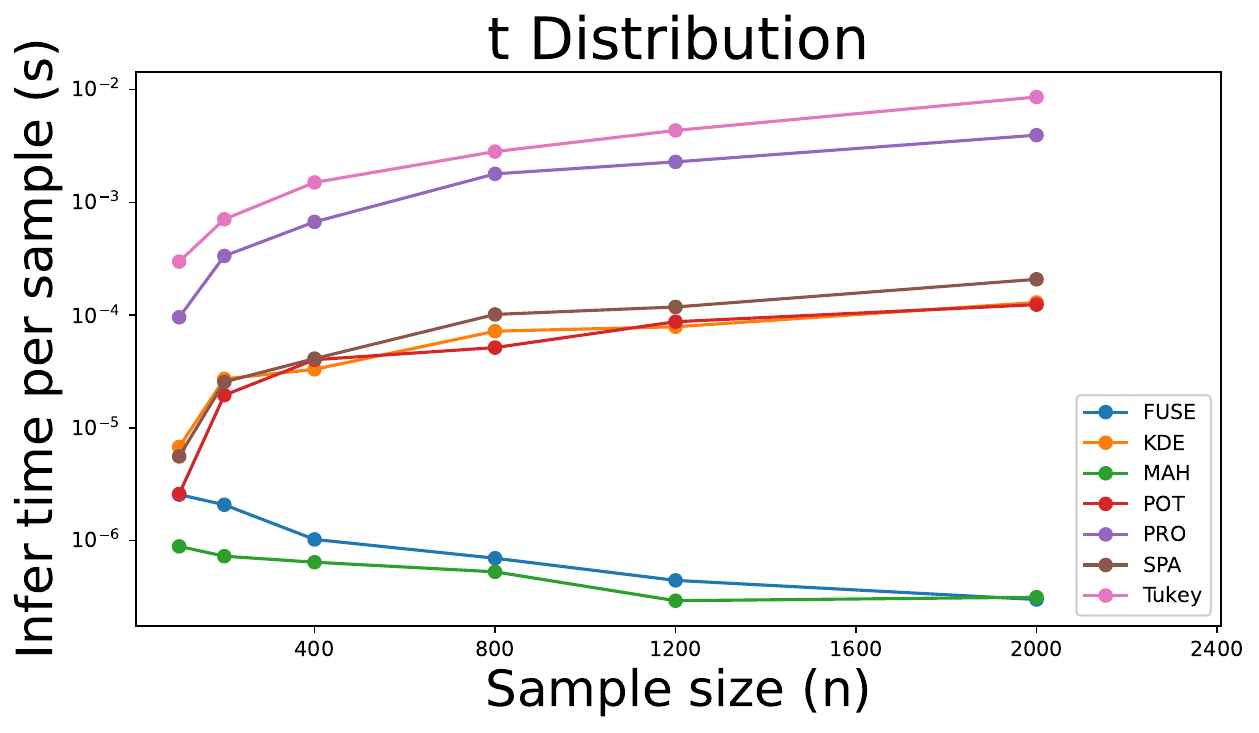}\\[4pt]
    \includegraphics[width=0.48\linewidth]{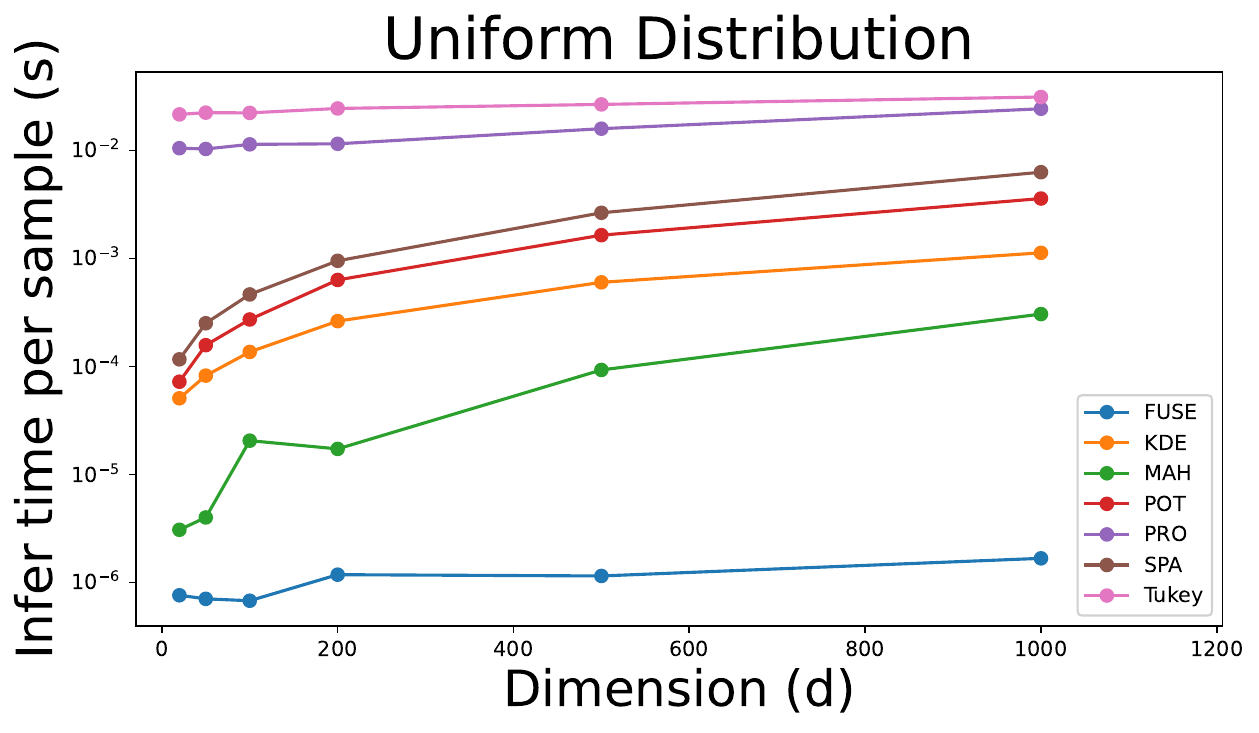}\hfill
    \includegraphics[width=0.48\linewidth]{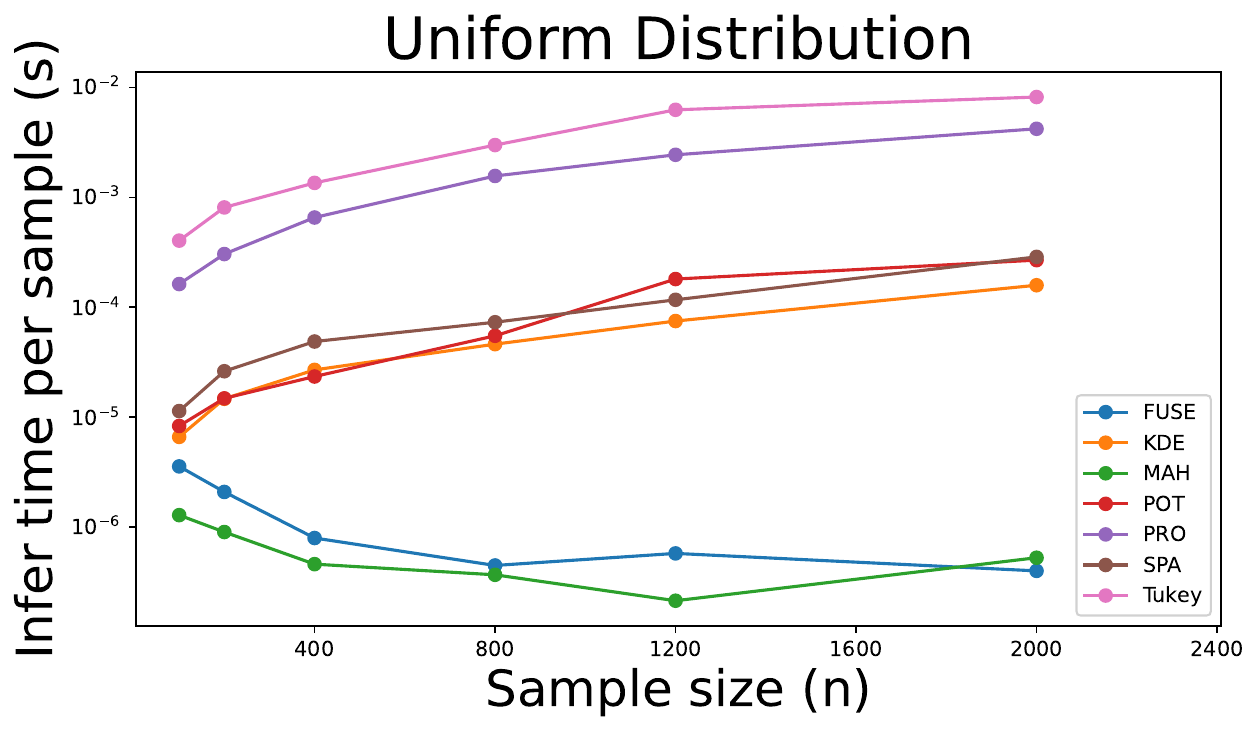}\\[4pt]
    \caption{Inference time per sample comparison of methods across increasing dimensions (left) and increasing sample sizes (right).}
    \label{fig:runtime_vs_dims_n}
\end{figure}

Inference for FUSE requires only a single forward pass, offering substantial advantages in efficiency over classical depth functions in both low and high dimensions. Figure~\ref{fig:runtime_vs_dims_n} reports the inference time per sample for all
methods on Gaussian, Student-$t$, and Uniform distributions, with FUSE evaluated using a single unified model where Global and Local are trained jointly. For each setting,
we measure how long it takes to compute scores for all $n$ data points and
then divide this total time by $n$. The curves show this average time per
sample. Across all settings, the inference time per sample of FUSE stays in the
microsecond range, about $10^{-6}$–$10^{-5}$ seconds, whereas classical depth
functions such as KDE, POT, PRO, SPA, and Tukey typically require
$10^{-4}$–$10^{-2}$ seconds per sample. Thus FUSE is roughly two to four
orders of magnitude faster at inference. When the sample size is fixed at
$n=500$ and the dimension increases, this gap widens further. When the
dimension is fixed at $d=5$ and the sample size increases, inference time per
sample grows for all classical methods, while FUSE and MAH remain nearly flat. Moreover, FUSE matches MAH in speed while being more robust and broadly applicable across data types.

\subsection{Real data exploration}
\label{sec:exp-real}

We next evaluate whether our centrality can reveal meaningful geometric structure in real, high-dimensional or non-Euclidean datasets.  
We consider three representative modalities, including images, time series and text, and study how the homotopy parameter $t\in[0,1]$ provides interpretable transitions between global and local views.  
These experiments complement the synthetic results and demonstrate that FUSE can serve as a simple, unified tool for exploring complex datasets.

All datasets follow a common process. We begin with either raw data or pretrained embeddings, followed by normalization when appropriate.  
From these representations, we compute a pairwise distance matrix using a task-suitable metric and train our neural centrality model based on this distance information. For visualization, we embed the learned centrality values into two dimensions using UMAP \citep{mcinnes2018umap} and examine how the resulting plots evolve as $t$ increases from $0$ to $1$.

Across all datasets, two consistent advantages emerge.  
First, FUSE highlights intrinsic organization within each modality, revealing clusters, gradients, and atypical samples that are difficult to identify in raw representation space.  
Second, the homotopy parameter $t$ provides a smooth and transparent way to shift emphasis from broad global structure ($t{=}0$) to fine-grained local density ($t{=}1$).  
Together, these findings suggest that FUSE provides an interpretable multi-scale summary of a dataset’s geometry.

\subsubsection{Images: CIFAR-10}
\label{sec:cifar_expt}

We first analyze the \texttt{airplane} class of CIFAR-10 \citep{krizhevsky2009learning}, which contains 60,000 color images across ten object classes.  
For visualization, we sample $n=1000$ airplane images and represent each image using a 512-dimensional CLIP embedding \citep{radford2021learning}.

All distances, graphs and baseline depth methods in this subsection are applied in the representation space. We focus on Global and classical depth-style baselines, including POT, PRO, MAH, SPA and Tukey, using the same configurations as in Section~\ref{sec:exp-synth} (see Supplement~\ref{app:otherdepth} for implementation details). We do not include KDE or Local here, because these methods are designed to highlight very local density peaks rather than to produce a single smooth center–outward ordering. On the CIFAR-10 CLIP graph, this leads to many sharp changes in their scores between nearby neighbors, so the graph-smoothness measure (introduced below) would mainly reflect this intended local variability instead of the quality of the method.

To assess how smoothly centrality varies across the data manifold, we construct a symmetrized 10-nearest-neighbor graph from the embeddings.  
Let $W$ be the adjacency matrix, and let $D$ be the diagonal degree matrix with $D_{ii} = \sum_j W_{ij}$. The graph Laplacian is defined as $L = D - W$.
Given centrality scores $s\in\mathbb{R}^n$, we evaluate smoothness using the normalized quadratic form  
$$
\frac{s^\top L s}{\|s\|_2^2}
= \frac{1}{2\|s\|_2^2}\sum_{i,j} W_{ij}(s_i-s_j)^2,
$$
which measures how much the scores vary across adjacent nodes \citep{dong2016learning}. Lower values indicate that neighboring points have more similar centrality, reflecting smoother structure.  
Because classical depth methods output scores on different scales, all methods use min–max normalization to $[0,1]$ before evaluation.

\begin{table}[!htbp]
\centering
\caption{Inference time per sample and graph smoothness on CIFAR-10 (\texttt{airplane}).}
\renewcommand{\arraystretch}{1.15}
\setlength{\tabcolsep}{4.5pt}
\begin{tabular*}{\linewidth}{@{\extracolsep{\fill}}lcccccc}
\toprule
\textbf{Metric} &
\textbf{Global} & \textbf{MAH} & \textbf{POT} & \textbf{PRO} & \textbf{SPA} & \textbf{Tukey} \\
\midrule
Infer time per sample (ms) &
0.003 & 0.224 & 18 & 17 & 96 & 25 \\
Graph Smoothness &
0.228 & 1.419 & 0.263 & 0.595 & 2.140 & 25.833 \\
\bottomrule
\end{tabular*}
\label{tab:depth_smoothness}
\end{table}

Table~\ref{tab:depth_smoothness} shows that Global achieves the lowest
graph-smoothness score (0.228) and the fastest inference time among all
methods, indicating the most coherent variation across the CLIP manifold.
POT attains a lower smoothness score than MAH, PRO, SPA, and Tukey, but is
significantly slower at inference. PRO and Tukey rely on approximate implementations and exhibit high
graph roughness on the CLIP manifold. For Tukey, this behavior matches its
known degeneracy in high dimensions. Its true population depth collapses onto
a nearly constant value for most points \citep{dutta2011some, yeon2025regularized},
so the remaining variation after min--max normalization is essentially
numerical noise rather than meaningful geometric structure. 
Overall, Global provides the best balance of interpretability, stability, and computational efficiency.

\begin{figure}[!htbp]
    \centering
    \includegraphics[width=1.0\linewidth]{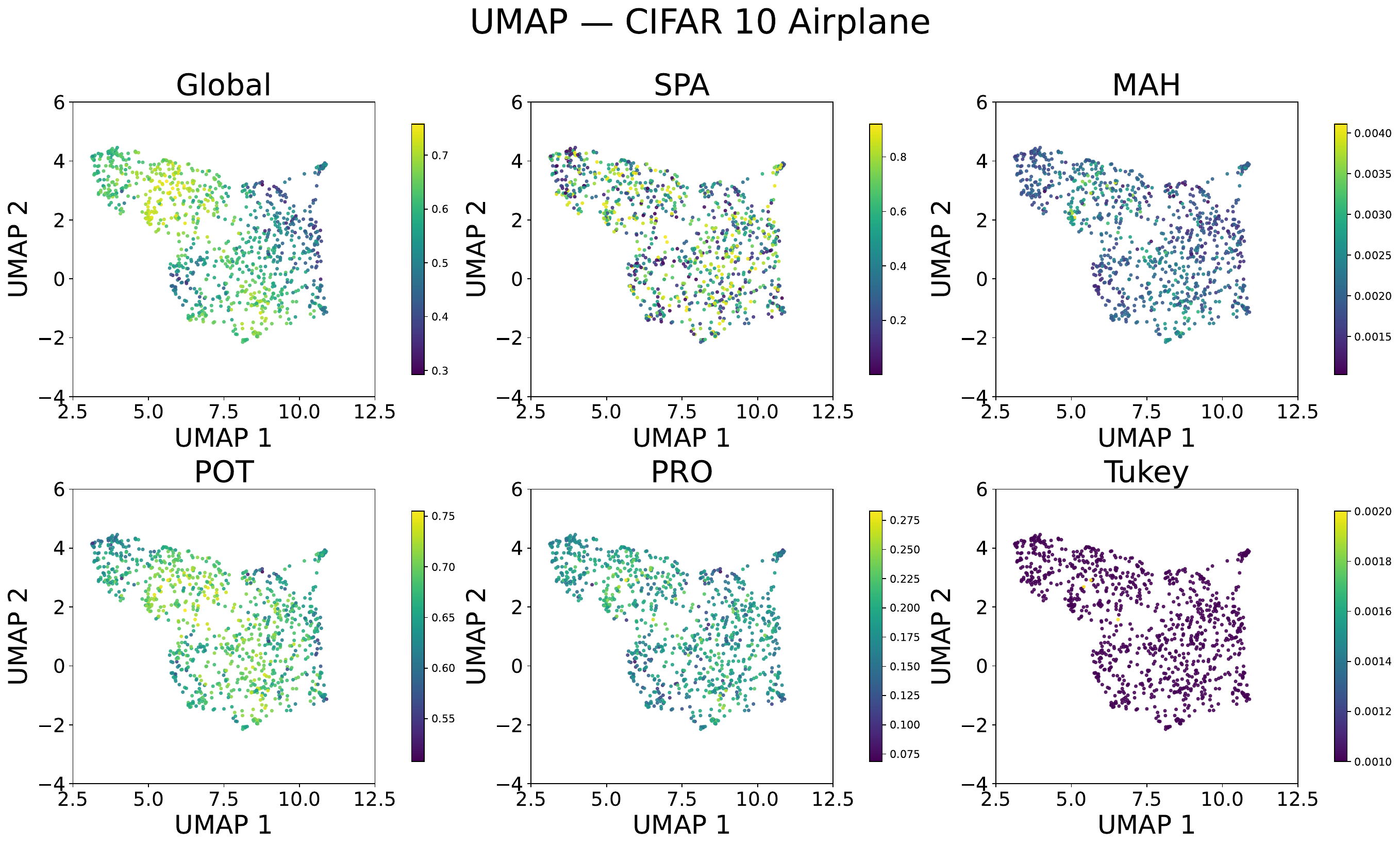}
    \caption{Comparison of centrality methods on CIFAR-10 (\texttt{airplane}).}
    \label{fig:cifar}
\end{figure}

Figure~\ref{fig:cifar} displays UMAP visualizations for all methods' results. Global produces scores that vary smoothly across the UMAP embedding. Nearby points have similar values and
there is a clear trend from low to high centrality regions.
POT is also relatively smooth while emphasizing several local
high scores rather than a single dominant center. SPA and PRO exhibit
more speckled, high-frequency variation, while MAH mainly reflects an approximately elliptical notion of centrality that does not match well the curved
shape of the embedding. Tukey depth is almost constant,
consistent with its high-dimensional degeneracy. The remaining variation appears
to be dominated by numerical noise rather than meaningful structures. 

\subsubsection{Images: MNIST}

We next examine MNIST \citep{lecun2002gradient}.  
Each grayscale $28\times28$ image is flattened to a 784-dimensional vector. We use a fixed subset of $10{,}000$ samples and compute their pairwise Euclidean distances for our model.

\begin{figure}[!htbp]
    \centering
    \includegraphics[width=\linewidth]{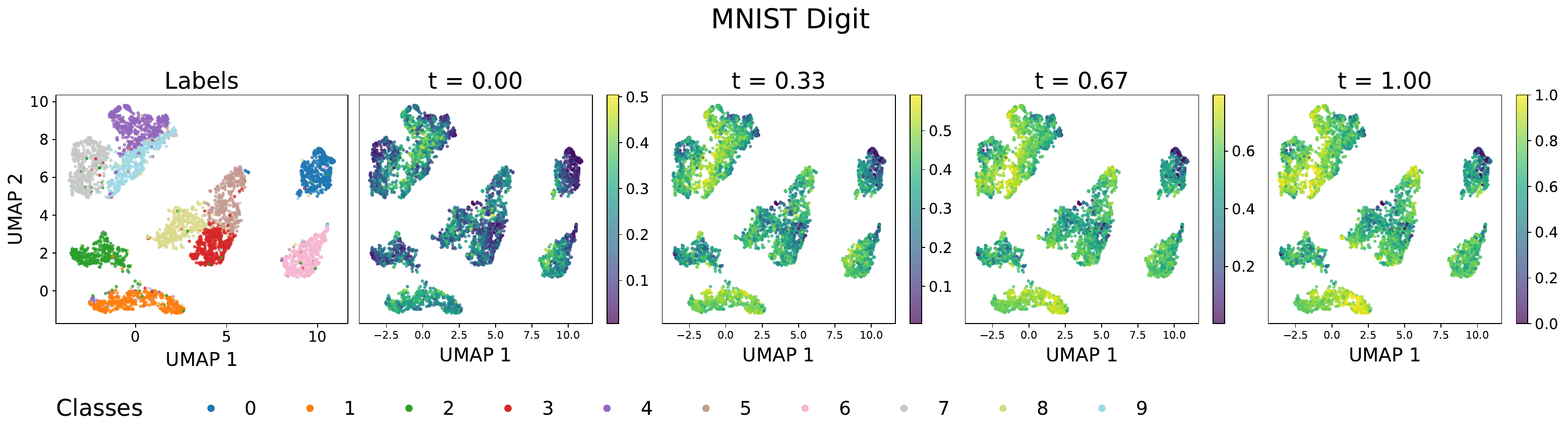}
    \caption{Homotopy centrality on MNIST.}
    \label{fig:mnist}
\end{figure}

Figure~\ref{fig:mnist} presents UMAP projections of the learned homotopy centrality.  
Clusters corresponding to the ten digits are visible even in the raw embedding, but FUSE provides a clearer interpretation.  
At $t{=}0$, global centrality captures inter-class structure. Digits such as 4, 7 and 9 overlap and exhibit relatively high centrality, while more distinctive digits, such as 0, 2 and 6, lie toward the periphery.  
As $t$ increases, FUSE places more emphasis on local density.  
At $t{=}1$, the highest centrality values concentrate around the dense cores of each digit cluster, providing a refined within-cluster perspective.

This evolution illustrates how homotopy centrality connects global geometric relationships with local density within each cluster, offering a unified multi-scale view of the dataset.

\subsubsection{Time series: FordA}

The FordA dataset \citep{dau2019ucr} contains one-dimensional time series labeled into two classes.  
We represent each series using 22-dimensional \texttt{catch22} features \citep{lubba2019catch22}, which preserve essential temporal characteristics while enabling efficient distance computation.  Euclidean distances in this feature space are used for training.

\begin{figure}[!htbp]
    \centering
    \includegraphics[width=1.0\linewidth]{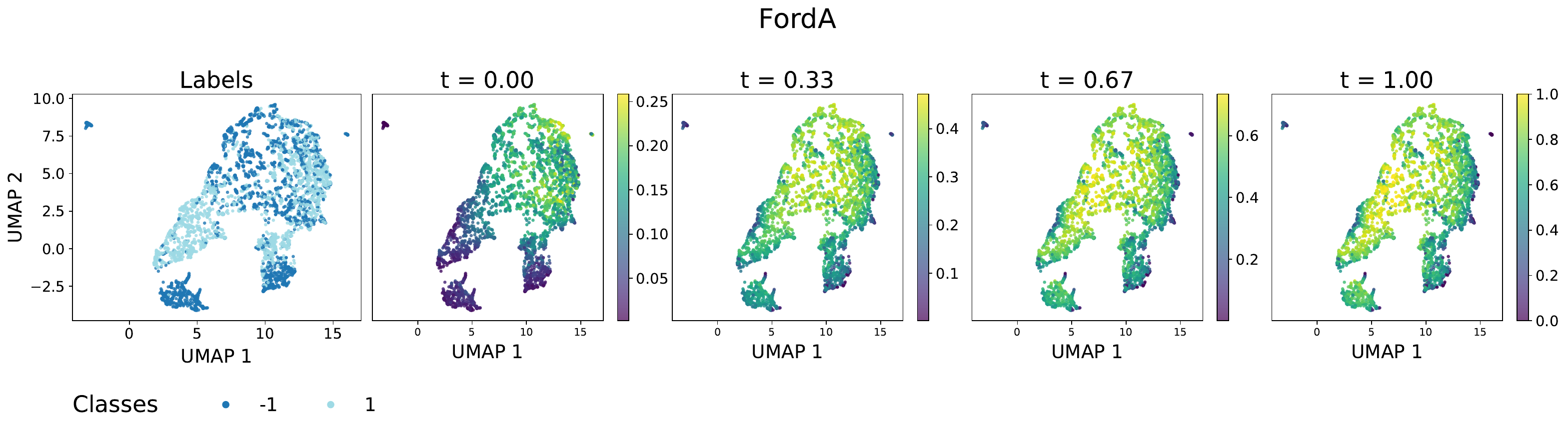} \\
    \caption{Homotopy centrality on FordA.}
    \label{fig:forda_depth}
\end{figure}

Figure~\ref{fig:forda_depth} shows the UMAP projection colored by homotopy centrality.  
Because the two classes strongly overlap, global centrality ($t{=}0$) is nearly flat, capturing only weak global structure.  
As $t$ increases, FUSE increasingly highlights local density variations. Dense clusters receive higher centrality, while scattered or boundary points receive lower values.  
By $t{=}1$, the local head captures boundaries that better align with class structure.

FordA demonstrates a scenario where local information is more informative than global geometry, and FUSE naturally adapts across this spectrum.

\subsubsection{Text: TweetEval (sentiment)}
The TweetEval benchmark \citep{barbieri2020tweeteval} contains Twitter datasets
for several text classification tasks. In our experiment, we focus on the
sentiment classification subset, which has three classes: negative (0), neutral
(1), and positive (2), and for computational efficiency we use a fixed subset of $1{,}000$ tweets. Each tweet is embedded using the pooled \texttt{[CLS]} output from \texttt{distilbert-base-uncased-finetuned-sst-2-english} \citep{sanh2019distilbert,socher2013recursive}.  
Pairwise cosine distances define the training metric.

\begin{figure}[!htbp]
    \centering
    \includegraphics[width=1.0\linewidth]{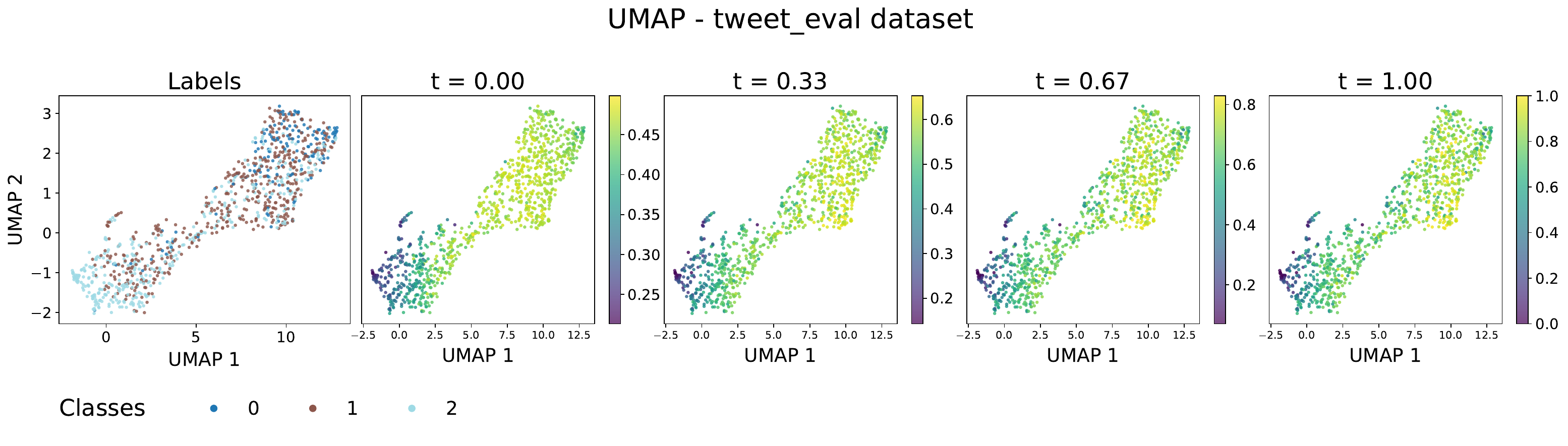}
    \caption{Homotopy centrality on TweetEval.}
    \label{fig:tweeteval}
\end{figure}

Figure~\ref{fig:tweeteval} shows the UMAP projection colored by homotopy centrality.  
Tweets with positive and negative sentiment lie at opposite ends, with neutral tweets forming a bridge connecting the two extremes.  
Both global and local scores assign higher centrality to this neutral region, reflecting its role as the manifold “middle”.

The dependence on $t$ is relatively mild for this dataset. Global centrality peaks in the neutral band, while local centrality slightly sharpens dense subregions and decreases scores for more diffuse boundary points. Across all settings, the structure remains coherent, indicating a smooth sentiment trend.

\subsection{Outlier detection}
\label{sec:outlier}
We further assess the effectiveness of FUSE for outlier detection in low and
moderately high-dimensional Euclidean data. We consider two benchmark datasets, Breastw and Ionosphere, drawn from the
ADBench and ODDS libraries with standard anomaly labels inherited from the original classification tasks. We use the standardized, preprocessed versions
provided by \citet{bouman2024unsupervised} in their public GitHub repository.\footnote{
\url{https://github.com/RoelBouman/outlierdetection}.} Following the
distinction between global and local density anomalies  as discussed in 
\citep{foorthuis2021nature,bouman2024unsupervised}, Breastw ($n=683$, $d=9$)
behaves as a global anomaly problem, since the anomalous classes are small but
clinically distinct, and tend to occupy regions of globally low density
relative to the main mass of normal samples. In contrast, Ionosphere
($n=351$, $d=33$) is a prototypical local anomaly problem, where
anomalous points lie in regions whose density is low compared to nearby normal
regions rather than forming a single, clearly separated peripheral cluster.

We compare three variants of our method, Global, Local, and the interpolated
family FUSE, with four standard one-class baselines, including Isolation Forest (IF;
\citealt{liu2008isolation}), Kernel Density Estimation (KDE;
\citealt{parzen1962estimation,silverman2018density}), Local Outlier Factor
(LOF; \citealt{breunig2000lof}), and One-Class SVM (OCSVM;
\citealt{scholkopf2001estimating}). For FUSE, we consider a grid of
homotopy parameters $t \in \{0.1, 0.2, \ldots, 0.9\}$ and treat $t$ as a
hyperparameter. Hence, in Table \ref{tab:od}, the single “FUSE’’ column reports the
cross-validated choice of $t$. All four baselines use the \texttt{scikit-learn} \citep{pedregosa2011scikit} implementations, and we tune
hyperparameters for both the baselines and our three variants by
cross-validation. Full grids and training details are given in
Supplement~\ref{app:outlier}.

\begin{table}[!htbp]
\centering
\caption{Outlier detection performance on Breastw and Ionosphere.}
\renewcommand{\arraystretch}{1.15}
\setlength{\tabcolsep}{5pt}
\begin{tabular*}{\linewidth}{@{\extracolsep{\fill}}llccccccc}
\toprule
\textbf{Dataset} & \textbf{Metric} &
\textbf{Global} & \textbf{Local} & \textbf{FUSE} &
\textbf{IF} & \textbf{LOF} & \textbf{OCSVM} & \textbf{KDE} \\
\midrule
\multirow{2}{*}{Breastw} 
 & AUC-ROC & 0.988 & 0.987 & 0.990 & 0.987 & 0.497 & 0.962 & 0.981 \\
 & AUC-PRC & 0.976 & 0.961 & 0.979 & 0.971 & 0.353 & 0.908 & 0.948 \\
\midrule
\multirow{2}{*}{Ionosphere} 
 & AUC-ROC & 0.870 & 0.959 & 0.958 & 0.853 & 0.903 & 0.764 & 0.942 \\
 & AUC-PRC & 0.875 & 0.950 & 0.950 & 0.810 & 0.859 & 0.773 & 0.938 \\
\bottomrule
\end{tabular*}
\label{tab:od}
\end{table}

Table~\ref{tab:od} reports both AUC-ROC and AUC-PRC for Breastw and
Ionosphere. On Breastw, where anomalies correspond to small but clinically
distinct diagnostic classes and global shifts dominate, Global already performs very strongly: it attains AUC-ROC~0.988 and AUC-PRC~0.976, outperforming all classical baselines. IF and KDE are the
strongest among the baselines, but still fall slightly short of Global in both
metrics, while LOF collapses on this dataset. FUSE provides a modest further
gain, achieving the best overall performance with AUC-ROC~0.990 and
AUC-PRC~0.979, slightly improving over both Global and Local.

Ionosphere presents a different challenge, with substantial overlap between
normal and anomalous samples and anomalies manifesting primarily as local
neighborhood irregularities. Here Global lags
behind density-based methods, with AUC-ROC~0.870 and AUC-PRC~0.875. In
contrast, Local and FUSE variants perform markedly
better. Local achieves AUC-ROC~0.959 and AUC-PRC~0.950, and FUSE attains
essentially the same values, both clearly outperforming Global and the
classical baselines. KDE and LOF are also strong on this dataset, but remain
below Local and FUSE on both metrics.

Overall, these results illustrate that the three FUSE variants capture
complementary aspects of abnormality. Global aligns well with datasets
dominated by broad distributional shifts (Breastw), Local is more
effective when anomalies are primarily local irregularities (Ionosphere), and
FUSE interpolates between them to provide robust performance without prior knowledge of whether a given dataset is ``global'' or ``local'' in nature.

\section{Conclusion and Discussion}
\label{sec：conclusion}
We presented FUSE, a neural framework for estimating centrality that combines
two complementary signals: a global ranking head capturing large-scale
geometric structure, and a local score-matching head capturing density-based
typicality. A homotopy interpolation links these two regimes, yielding a
continuum of centrality scores that trade off global robustness and local
sensitivity. This unified design enables FUSE to operate across heterogeneous
data types and to provide interpretable, scalable centrality estimates in
high-dimensional and non-Euclidean settings.

From a practical standpoint, FUSE is best viewed as a lightweight module
operating on fixed representations, with a simple trainable architecture. In our experiments we use a shared encoder consisting
of two fully connected layers with GELU activations \citep{hendrycks2016gaussian} and width $q$, followed
by linear heads for the global and local scores. Choosing $q \in \{32,64,128\}$ already yields strong performance across all of our
benchmarks. For more complex data, practitioners can increase the width or
depth of this encoder, or replace it with a domain-specific neural network,
while keeping the FUSE heads and loss functions unchanged.

Future directions include exploring alternative similarity notions, extending
the framework to more structured or multimodal data, and developing a more
systematic understanding of how to adapt the noise parameter
$\eta$ for arbitrary data types. Overall, we hope FUSE provides a flexible
foundation for incorporating depth and centrality ideas into modern neural
models and for developing more interpretable, geometry-aware learning
systems.




\vskip 0.2in
\bibliography{JMLRsample}

\newpage

\appendix

\renewcommand{\thetable}{S\arabic{table}}
\renewcommand{\thefigure}{S\arabic{figure}}
\renewcommand{\theequation}
{S\arabic{equation}}
\renewcommand{\appendixname}{} 
\renewcommand{\thesection}{S\arabic{section}}
\renewcommand{\thesubsection}{S\arabic{section}.\arabic{subsection}}

\setcounter{equation}{0}
\setcounter{table}{0}
\setcounter{figure}{0}
\begin{center}
	{\bf  \Large
		Supplementary Material for ``A Trainable Centrality Framework for Modern Data''}  \\
\end{center}

\section{Neural architecture for FUSE}
\label{app:net-arch}

For all experiments, the trainable part of FUSE follows a common pipeline.
Starting from the fixed representations $Z_i = \psi(X_i)$, we apply a shared
encoder $e_{\theta_e} : \mathbb{R}^p \to \mathbb{R}^{q}$ implemented as
two fully connected layers with GELU \citep{hendrycks2016gaussian} activations, followed by a linear
projection to a $q$-dimensional feature.  This shared feature is then passed
to two scalar linear heads: the global ranking head $g_\theta$ and the local
DSM head $l_\theta$, as described in Section~\ref{sec:method}.  In all our
experiments, the hidden width of the encoder and the projection dimension $q$
are taken to be the same (e.g., $q \in \{32,64,128\}$ depending on the setting).

The only architectural difference across experimental settings is this hidden
width, which is chosen to reflect the complexity and dimensionality of the
inputs:
\begin{itemize}
  \item \textbf{Synthetic data (Section~\ref{sec:exp-synth}).}
  The shared encoder uses width $(32,32)$ for the two hidden layers, followed
  by a $32$-dimensional linear projection (so $q=32$). These low-dimensional
  toy examples mainly serve to validate qualitative depth-like behavior, so a
  small network suffices.

  \item \textbf{Real-data visualizations (Section~\ref{sec:exp-real}).}
  For image, time series, and text experiments based on pretrained embeddings,
  we use width $(64,64)$ and a $64$-dimensional projection (so $q=64$). The
  representations are moderately high-dimensional but already informative, so
  we use a slightly larger encoder while avoiding overparameterization.

  \item \textbf{Outlier detection (Section~\ref{sec:outlier}).}
  For the ODDS datasets, we operate directly on raw feature vectors. To capture
  more complex structure in these inputs, we use two hidden layers of size
  $(128,128)$ and a $128$-dimensional projection (so $q=128$) in the shared
  encoder.
\end{itemize}

In all experiments, the number of anchor points per global pair is fixed at
$64$, and each data point is perturbed $8$ times per epoch for the DSM
objective.  All models are trained for $30$ epochs using the Adam optimizer
(learning rate $10^{-3}$, weight decay $0.0$, batch size $128$).  GELU
activations are used between hidden layers, no dropout is applied, and the
heads $g_\theta$ and $l_\theta$ are implemented as single-layer linear maps on
top of the shared encoder output.

\section{Different sampling strategy investigation}
\label{sec:diff,sample}

We evaluate four global sampling schemes for constructing triplets $(S_0,S_1,S_2)$ used to train the global head.  
Throughout this section, let the training sample be $\{X_i\}_{i=1}^n$.  
Each scheme, denoted Schemes 1–4, specifies how anchors and candidate pairs are drawn from $\{X_i\}_{i=1}^n$ in order to estimate the pairwise ranking probabilities $\hat{\pi}(X_i,X_j)$ defined in Section~\ref{sec:global}.

\paragraph{Scheme 1 (default).}
We randomly partition the sample into three disjoint subsets of (approximately) equal size,
$$
S_0 \cup S_1 \cup S_2 = \{X_i\}_{i=1}^n,\qquad S_a \cap S_b = \emptyset \ \text{for } a\neq b \in \{0,1,2\}.
$$
We then form $\lfloor n/3 \rfloor$ candidate pairs
$(X_{i}, X_{j})$ with $X_{i}\in S_1$ and $X_{j}\in S_2$ by matching elements by index, and, for each pair, sample a small number (64 in the experiments) of anchors from $S_0$ to estimate $\hat{\pi}(X_i,X_j)$.  
This scheme provides a good balance between efficiency and accuracy and is used as the default configuration in all main experiments.

\paragraph{Scheme 2.}
We use the same partitioning as in Scheme~1, but instead of index-wise matching, we form all possible pairs between $S_1$ and $S_2$, i.e., the Cartesian product $S_1\times S_2$, which yields $O(n^2)$ candidate pairs.  
Anchors are still subsampled from $S_0$ for each pair.

\paragraph{Scheme 3.}
This scheme is identical to Scheme~2 in how candidate pairs are formed, but all anchors in $S_0$ are used when estimating $\hat{\pi}(X_i,X_j)$ (no subsampling).  
It therefore corresponds to an exhaustive anchor-selection strategy.

\paragraph{Scheme 4.}
The most exhaustive variant sets
$$
S_0 = S_1 = S_2 = \{X_i\}_{i=1}^n,
$$
so that anchors and candidates can be any points in the sample.  
This produces $O(n^3)$ triplets $(X_{i_0},X_{i_1},X_{i_2})$ and is primarily of conceptual interest due to its high computational cost.

\paragraph{Empirical comparison.}
We compare all four schemes under the same experimental settings as in the main paper, covering both synthetic data and outlier detection tasks.  
All models use identical hyperparameters. Schemes~1 and~2 sample 64 anchors per candidate pair, while Schemes~3 and~4 use all available anchors.

Figures~\ref{fig:app,gaussian}--\ref{fig:app,mixture} and Tables~\ref{tab:normal_results}--\ref{tab:uniform_results} report the synthetic results.  
The design follows Section~\ref{sec:exp-synth}, but with $1000$ (rather than $5000$) two-dimensional samples in each setting to reduce computational cost.  
Figures~\ref{fig:app,gaussian}--\ref{fig:app,mixture} visualize the homotopy centrality contours across different distributions and sampling schemes, while Tables~\ref{tab:normal_results}--\ref{tab:uniform_results} summarize Spearman’s and Kendall’s rank correlations on the test set, as well as model training time and average inference time per sample.  
Tables~\ref{tab:breastw_results}--\ref{tab:ionosphere_results} report the
detailed results for the outlier detection experiments in
Section~\ref{sec:outlier}. The network architecture follows
Supplement~\ref{app:net-arch}, and here we use a fixed configuration with two hidden layers of size $(128,128)$. In contrast to
Section~\ref{sec:outlier}, where hyperparameters are tuned by
cross-validation, all parameters are kept fixed in this supplement to reduce computational cost. These tables therefore illustrate the behavior of FUSE under a simple default architecture, while the main results in Section~\ref{sec:outlier}
correspond to tuned models.

Across all datasets, the four schemes yield very similar rank correlations and detection accuracy. However, Scheme~1 achieves this performance with substantially lower computational and memory cost. We therefore adopt Scheme~1 as the default global sampling configuration in all experiments reported in the paper.

\begin{figure}[!htbp]
    \centering
    \includegraphics[width=1.0\linewidth]{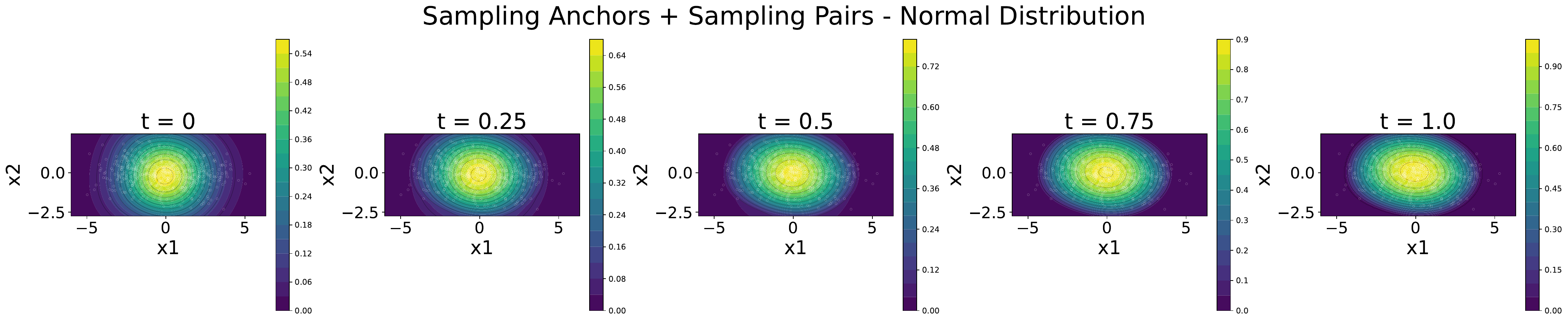}\\[2pt]
    \includegraphics[width=1.0\linewidth]{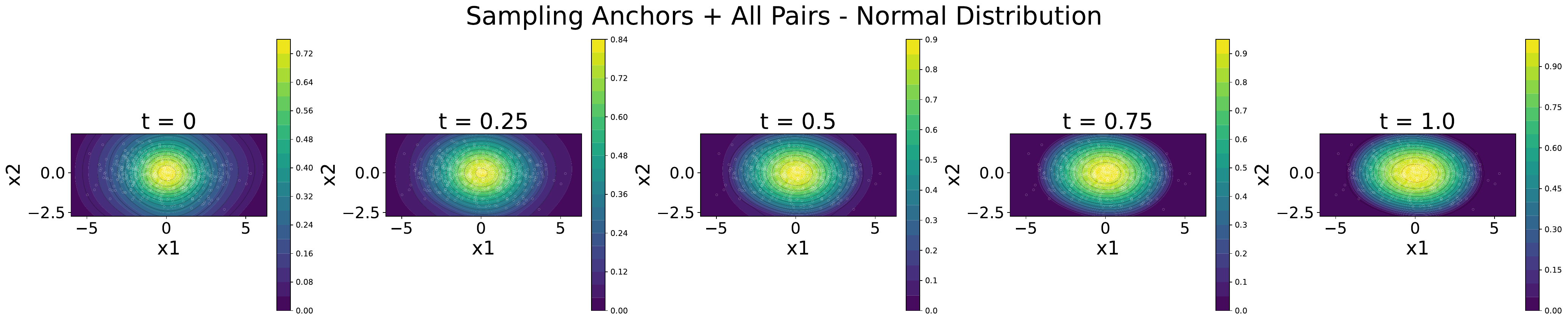}\\[2pt]
    \includegraphics[width=1.0\linewidth]{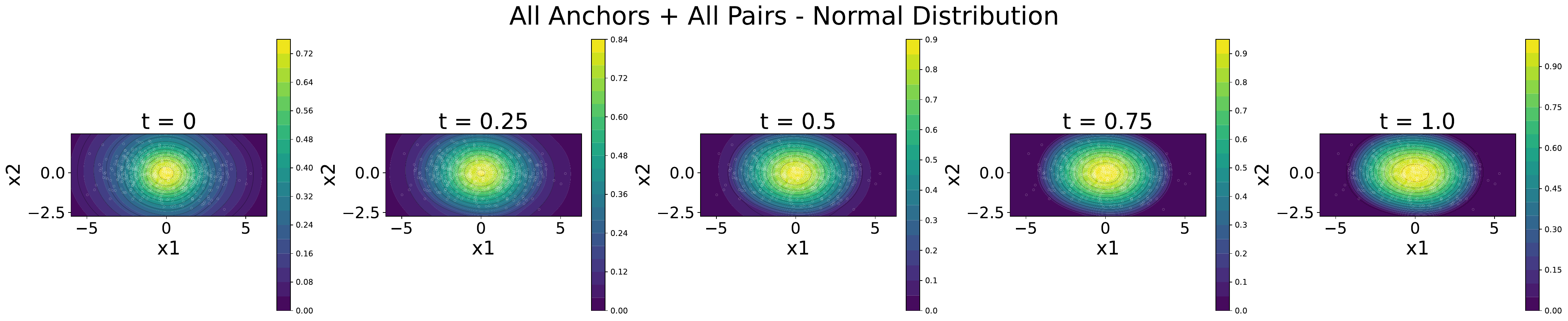}\\[2pt]
    \includegraphics[width=1.0\linewidth]{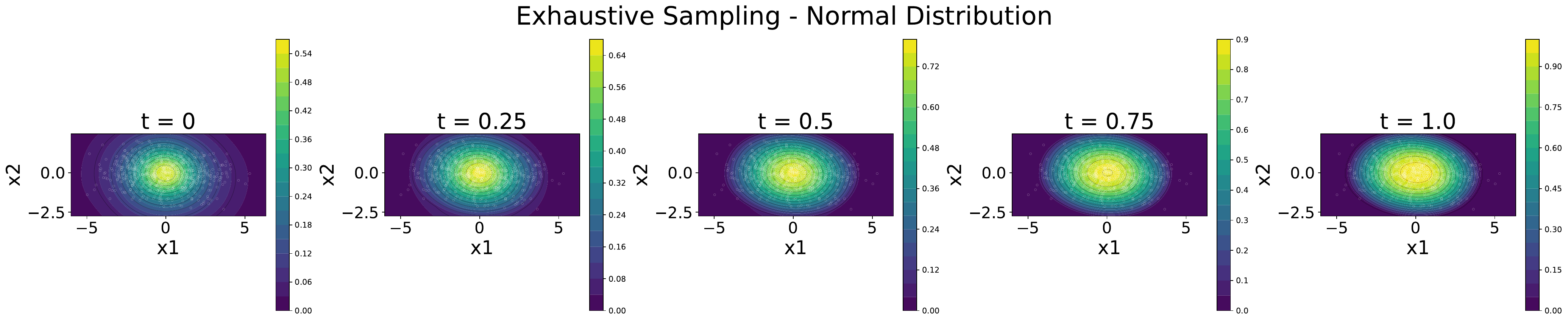}
    \caption{Homotopy centrality contours on Normal distribution.}    \label{fig:app,gaussian}
\end{figure}

\begin{figure}[!htbp]
    \centering
    \includegraphics[width=1.0\linewidth]{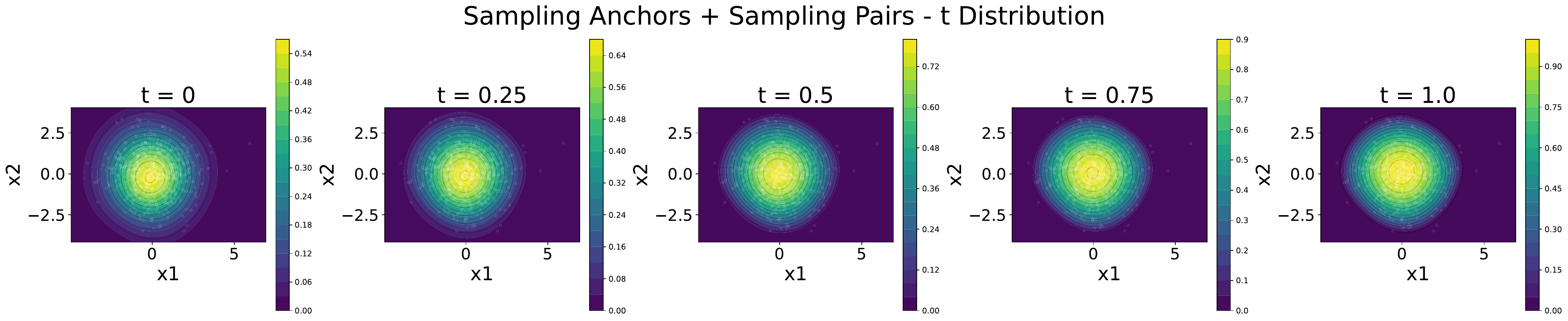}\\[2pt]
    \includegraphics[width=1.0\linewidth]{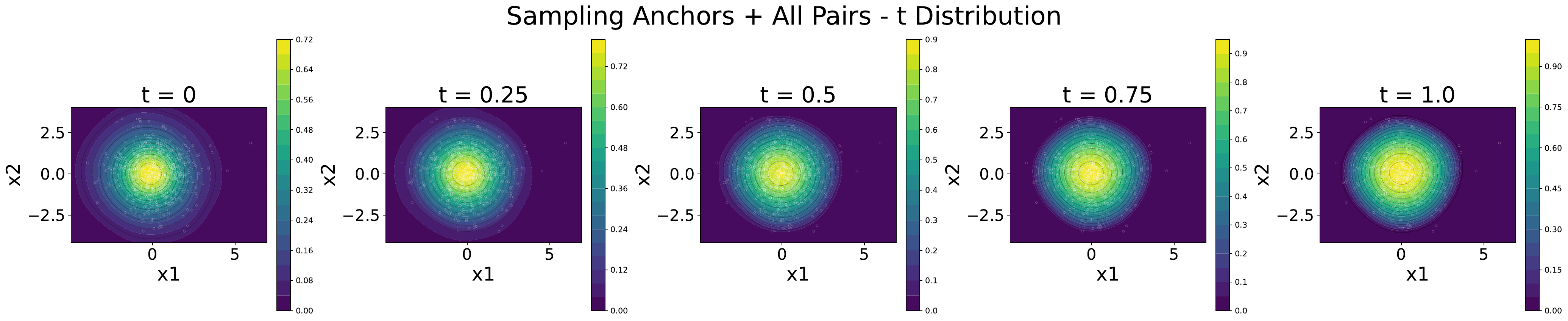}\\[2pt]
    \includegraphics[width=1.0\linewidth]{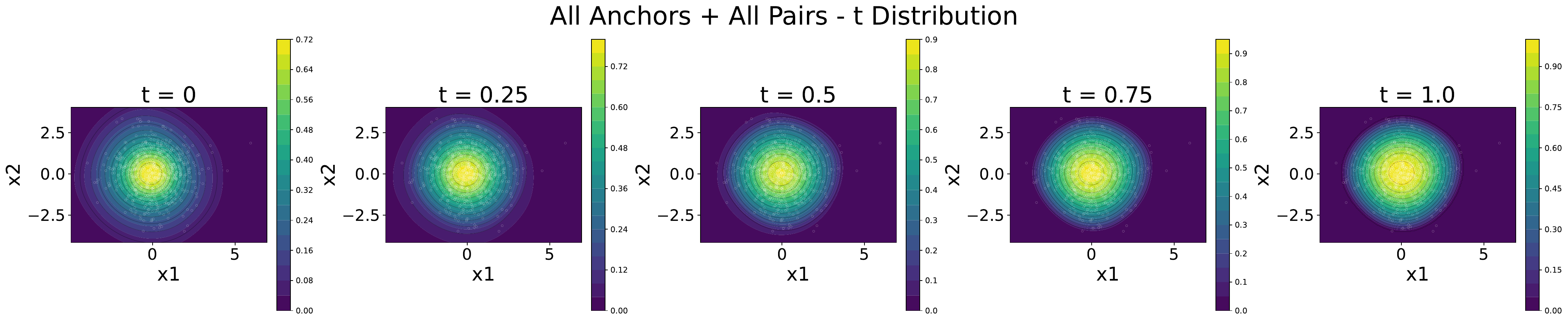}\\[2pt]
    \includegraphics[width=1.0\linewidth]{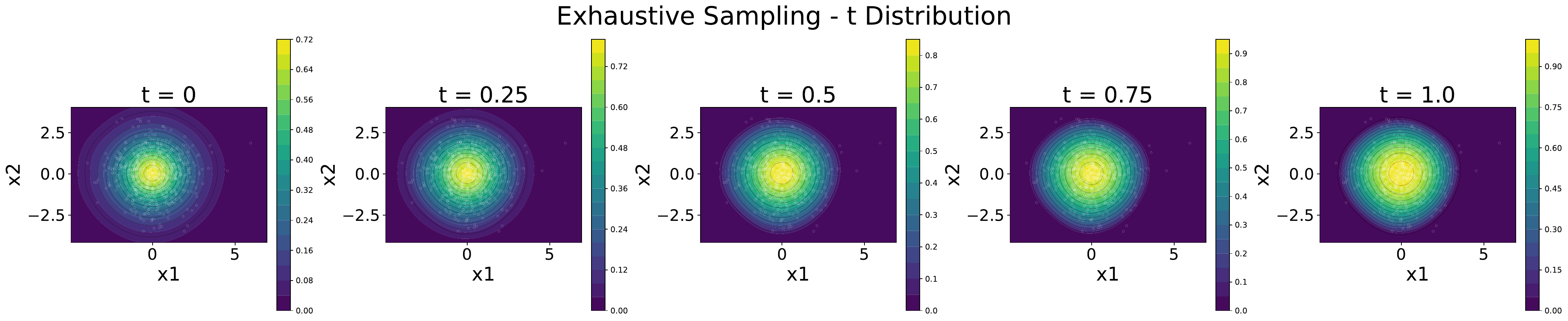}
    \caption{Homotopy centrality contours on Student-$t_{10}$ distribution.} 
    
\end{figure}

\begin{figure}[!htbp]
    \centering
    \includegraphics[width=1.0\linewidth]{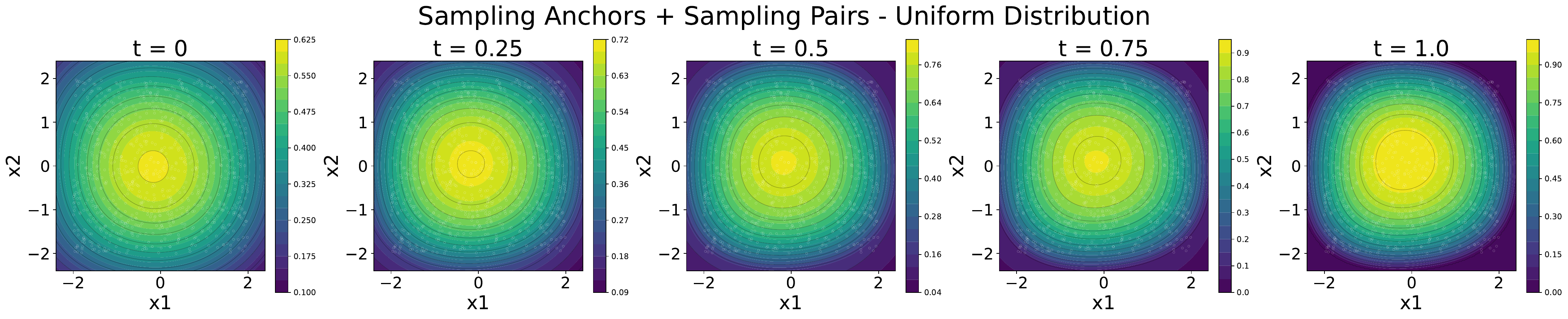}\\[2pt]
    \includegraphics[width=1.0\linewidth]{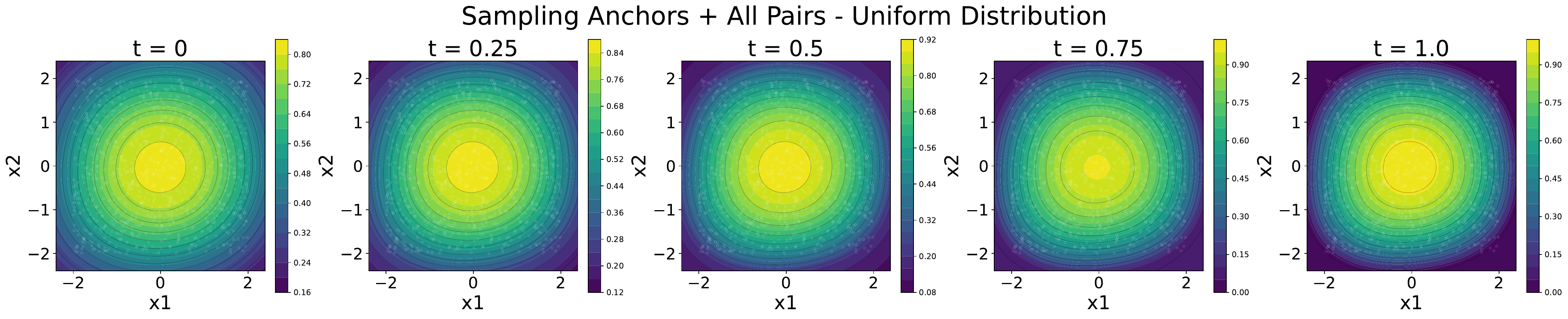}\\[2pt]
    \includegraphics[width=1.0\linewidth]{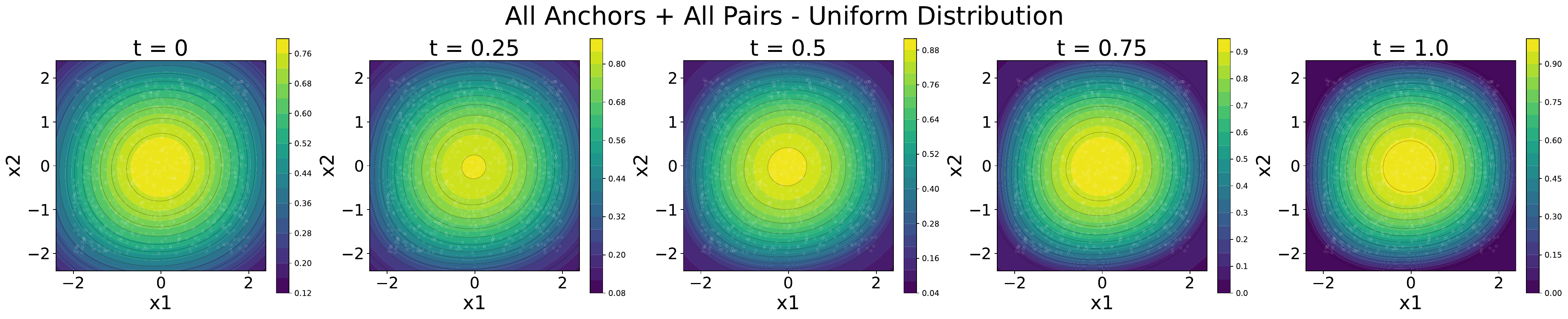}\\[2pt]
    \includegraphics[width=1.0\linewidth]{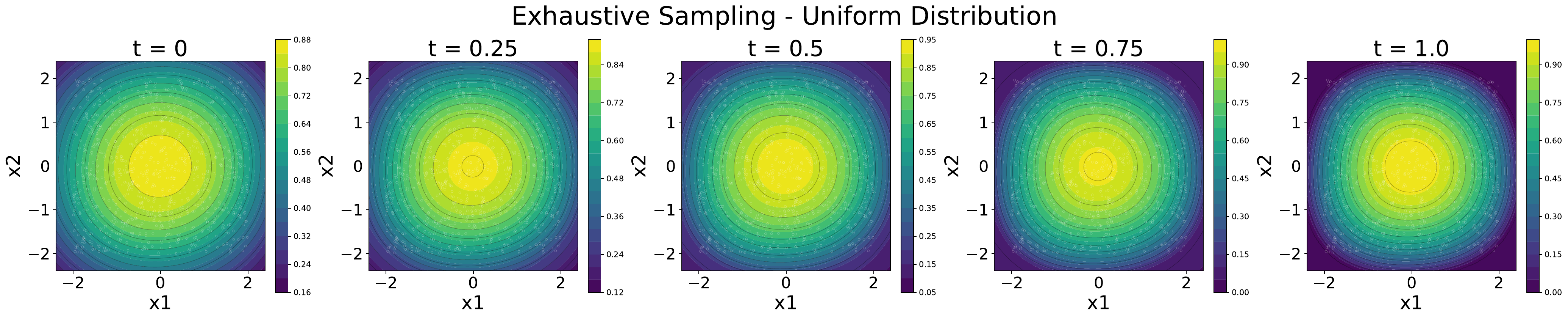}
    \caption{Homotopy centrality contours on Uniform distribution.} 
\end{figure}

\begin{figure}[!htbp]
    \centering
    \includegraphics[width=1.0\linewidth]{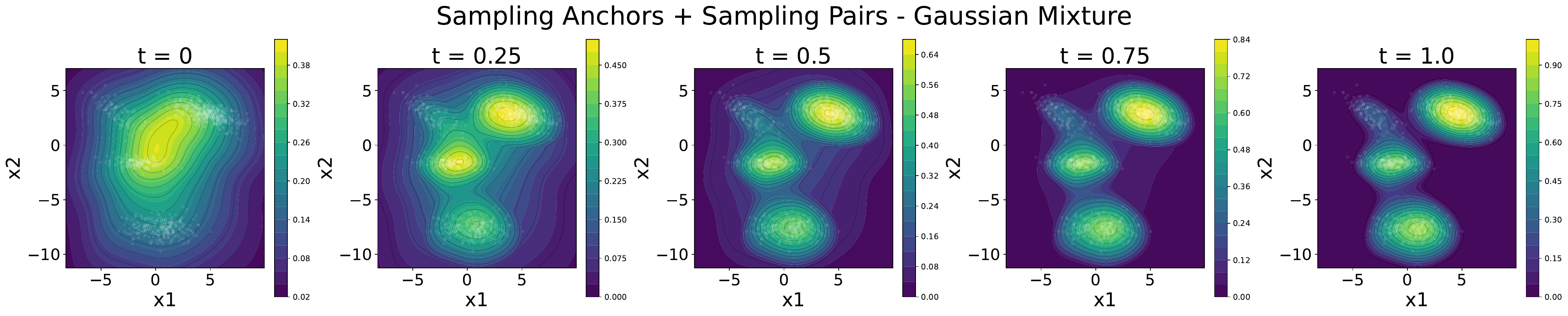}\\[2pt]
    \includegraphics[width=1.0\linewidth]{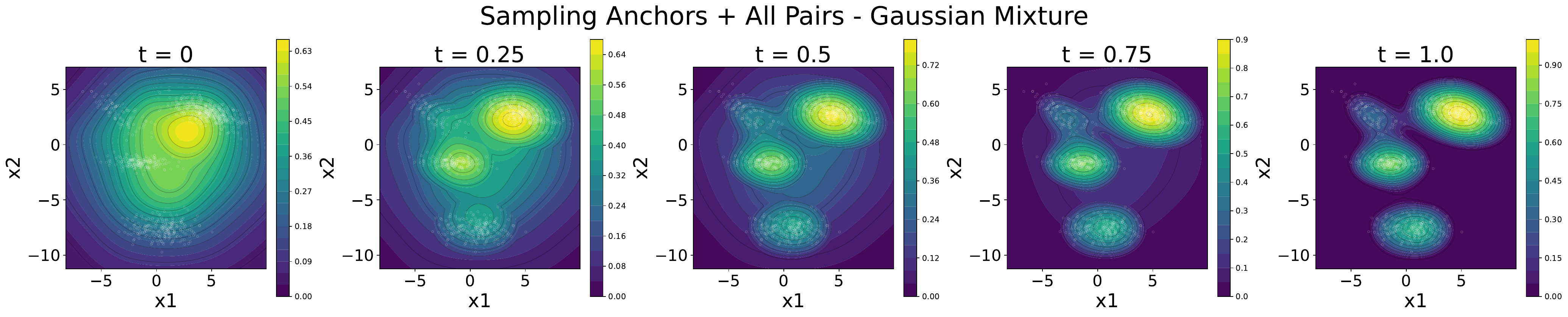}\\[2pt]
    \includegraphics[width=1.0\linewidth]{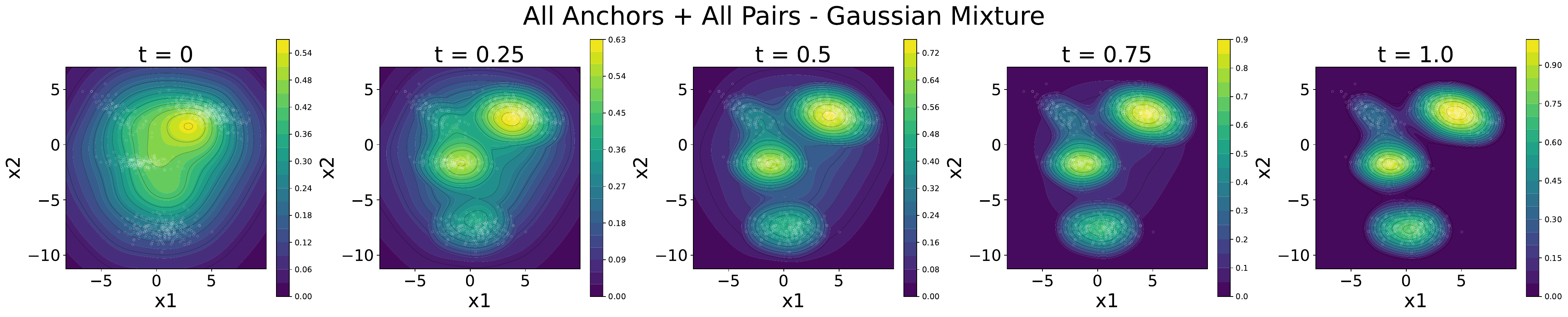}\\[2pt]
    \includegraphics[width=1.0\linewidth]{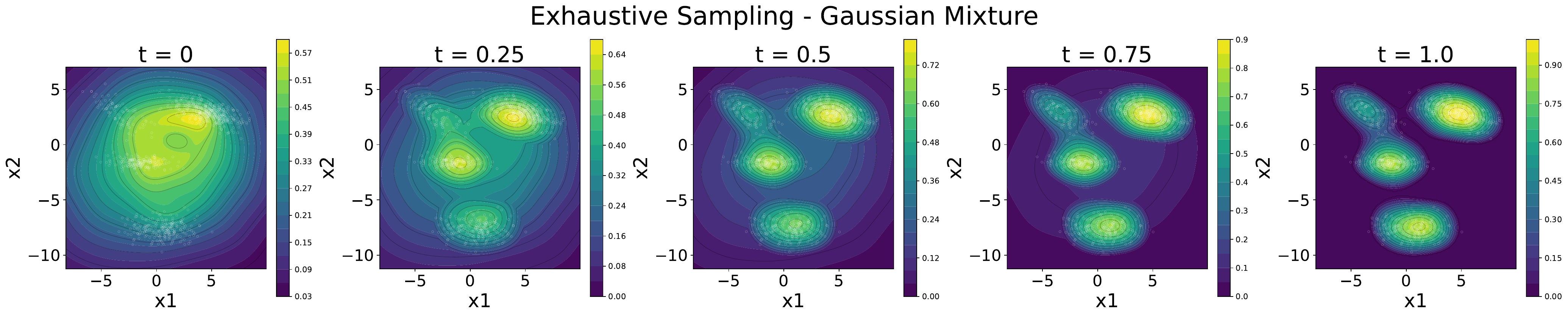}
    \caption{Homotopy centrality contours on Gaussian mixture.}     \label{fig:app,mixture}
\end{figure}

\begin{table}[!htbp]
\centering
\caption{Spearman’s $\rho$, Kendall’s $\tau$, training time, and average inference time per sample on the Normal distribution.}
\renewcommand{\arraystretch}{1.15}
\setlength{\tabcolsep}{5pt}
\begin{tabular*}{\linewidth}{@{\extracolsep{\fill}}lcccc}
\toprule
\textbf{Algorithm} &
\textbf{Spearman} &
\textbf{Kendall} &
\textbf{Train Time (s)} &
\textbf{Infer Time (ms)} \\
\midrule
Scheme~1 (Global) & 0.9887 & 0.9095 & 4.81 & 0.369 \\
Scheme~1 (Local)  & 0.9778 & 0.8807 & 4.81 & 0.369 \\
Scheme~2 (Global) & 0.9899 & 0.9168 & 137.08 & 0.379 \\
Scheme~2 (Local)  & 0.9758 & 0.8773 & 137.08 & 0.379 \\
Scheme~3 (Global) & 0.9909 & 0.9216 & 101.68 & 0.478 \\
Scheme~3 (Local)  & 0.9799 & 0.8895 & 101.68 & 0.478 \\
Scheme~4 (Global) & 0.9928 & 0.9353 & 609.46 & 0.420 \\
Scheme~4 (Local)  & 0.9732 & 0.8712 & 609.46 & 0.420 \\
\bottomrule
\end{tabular*}
\label{tab:normal_results}
\end{table}

\begin{table}[!htbp]
\centering
\caption{Spearman’s $\rho$, Kendall’s $\tau$, training time, and average inference time per sample on the Student-$t_{10}$ distribution.}
\renewcommand{\arraystretch}{1.15}
\setlength{\tabcolsep}{5pt}
\begin{tabular*}{\linewidth}{@{\extracolsep{\fill}}lcccc}
\toprule
\textbf{Algorithm} &
\textbf{Spearman} &
\textbf{Kendall} &
\textbf{Train Time (s)} &
\textbf{Infer Time (ms)} \\
\midrule
Scheme~1 (Global) & 0.9953 & 0.9421 & 4.78 & 0.349 \\
Scheme~1 (Local)  & 0.9986 & 0.9684 & 4.78 & 0.349 \\
Scheme~2 (Global) & 0.9941 & 0.9343 & 158.86 & 0.476 \\
Scheme~2 (Local)  & 0.9972 & 0.9546 & 158.86 & 0.476 \\
Scheme~3 (Global) & 0.9943 & 0.9355 & 127.09 & 0.485 \\
Scheme~3 (Local)  & 0.9989 & 0.9720 & 127.09 & 0.485 \\
Scheme~4 (Global) & 0.9990 & 0.9740 & 569.43 & 0.607 \\
Scheme~4 (Local)  & 0.9984 & 0.9670 & 569.43 & 0.607 \\
\bottomrule
\end{tabular*}
\label{tab:t_results}
\end{table}

\begin{table}[!htbp]
\centering
\caption{Spearman’s $\rho$, Kendall’s $\tau$, training time, and average inference time per sample on the Uniform distribution.}
\renewcommand{\arraystretch}{1.15}
\setlength{\tabcolsep}{5pt}
\begin{tabular*}{\linewidth}{@{\extracolsep{\fill}}lcccc}
\toprule
\textbf{Algorithm} &
\textbf{Spearman} &
\textbf{Kendall} &
\textbf{Train Time (s)} &
\textbf{Infer Time (ms)} \\
\midrule
Scheme~1 (Global) & 0.9957 & 0.9460 & 4.86 & 0.506 \\
Scheme~1 (Local)  & 0.9877 & 0.9032 & 4.86 & 0.506 \\
Scheme~2 (Global) & 0.9926 & 0.9281 & 137.73 & 0.581 \\
Scheme~2 (Local)  & 0.9718 & 0.8523 & 137.73 & 0.581 \\
Scheme~3 (Global) & 0.9968 & 0.9524 & 100.51 & 0.543 \\
Scheme~3 (Local)  & 0.9891 & 0.9115 & 100.51 & 0.543 \\
Scheme~4 (Global) & 0.9922 & 0.9272 & 589.98 & 0.468 \\
Scheme~4 (Local)  & 0.9851 & 0.8994 & 589.98 & 0.468 \\
\bottomrule
\end{tabular*}
\label{tab:uniform_results}
\end{table}

\begin{table}[!htbp]
\centering
\caption{Test AUC-ROC, AUC-PRC, training time, and average inference time per sample for the Breastw dataset.}
\renewcommand{\arraystretch}{1.15}
\setlength{\tabcolsep}{5pt}
\begin{tabular*}{\linewidth}{@{\extracolsep{\fill}}lcccc}
\toprule
\textbf{Model} &
\textbf{AUC-ROC} &
\textbf{AUC-PRC} &
\textbf{Train Time (s)} &
\textbf{Infer Time (ms)} \\
\midrule
Scheme~1 (Global) & 0.950 & 0.912 & 4.227   & 1.276 \\
Scheme~1 (Local)  & 0.983 & 0.953 & 4.227   & 1.276 \\

Scheme~2 (Global) & 0.945 & 0.906 & 18.056 & 1.132 \\
Scheme~2 (Local)  & 0.983 & 0.955 & 18.056 & 1.132 \\

Scheme~3 (Global) & 0.948 & 0.902 & 14.358 & 1.590 \\
Scheme~3 (Local)  & 0.984 & 0.955 & 14.358 & 1.590 \\

Scheme~4 (Global) & 0.954 & 0.931 & 92.718 & 1.336 \\
Scheme~4 (Local)  & 0.985 & 0.955 & 92.718 & 1.336 \\
\bottomrule
\end{tabular*}
\label{tab:breastw_results}
\end{table}

\begin{table}[!htbp]
\centering
\caption{Test AUC-ROC, AUC-PRC, training time, and average inference time per sample for Ionosphere dataset.}
\renewcommand{\arraystretch}{1.15}
\setlength{\tabcolsep}{5pt}
\begin{tabular*}{\linewidth}{@{\extracolsep{\fill}}lcccc}
\toprule
\textbf{Model} &
\textbf{AUC-ROC} &
\textbf{AUC-PRC} &
\textbf{Train Time (s)} &
\textbf{Infer Time (ms)} \\
\midrule
Scheme~1 (Global) & 0.758 & 0.748 & 0.961 & 0.653 \\
Scheme~1 (Local)  & 0.891 & 0.854 & 0.961 & 0.653 \\
Scheme~2 (Global) & 0.779 & 0.744 & 3.873 & 0.535 \\
Scheme~2 (Local)  & 0.865 & 0.818 & 3.873 & 0.535 \\
Scheme~3 (Global) & 0.742 & 0.714 & 3.477 & 0.767 \\
Scheme~3 (Local)  & 0.862 & 0.818 & 3.477 & 0.767 \\
Scheme~4 (Global) & 0.757 & 0.719 & 19.312 & 0.937 \\
Scheme~4 (Local)  & 0.859 & 0.806 & 19.312 & 0.937 \\
\bottomrule
\end{tabular*}
\label{tab:ionosphere_results}
\end{table}

\section{Depth function implementations}
\label{app:otherdepth}

For benchmarking, we re-implemented several classical depth functions in Python, following the formulations in the \texttt{data-depth} package.\footnote{\url{https://github.com/data-depth/library}}  
Let the reference sample be $\{X_i\}_{i=1}^n \subset \mathbb{R}^d$, and let $X_{1:n} \in \mathbb{R}^{n\times d}$ denote the corresponding data matrix with rows $X_i^\top$.  
Each depth function takes $X_{1:n}$ as reference and evaluates a depth value for a query point $x \in \mathbb{R}^d$.

\paragraph{Mahalanobis depth.} \citep{mahalanobis1936}  
Let the sample mean and the sample variance be
$$
\bar{X} = \frac{1}{n}\sum_{i=1}^n X_i, \quad  \hat{\Sigma} = \frac{1}{n-1}\sum_{i=1}^n (X_i - \bar{X})(X_i - \bar{X})^\top.
$$
For numerical stability, we add a small ridge $\varepsilon I_d$ with $\varepsilon = 10^{-9}$.  
The regularized Mahalanobis depth is
$$
D_{\mathrm{mah}}(x \mid X_{1:n})
= \frac{1}{1 + (x-\bar{X})^\top (\hat{\Sigma} + \varepsilon I_d)^{-1}(x-\bar{X})}.
$$
Instead of explicitly inverting $\hat{\Sigma}$, we compute a Cholesky factorization
$L L^\top = \hat{\Sigma} + \varepsilon I_d$ and evaluate the quadratic form as
$\|L^{-1}(x-\bar{X})\|_2^2$ for efficiency.

\paragraph{Spatial depth.} \citep{serfling2002depth}  
Define the spatial sign function $s(v)$ by
$$
s(v) =
\begin{cases}
v / \|v\|_2, & v \neq 0,\\[3pt]
0, & v = 0.
\end{cases}
$$
For affine invariance, we whiten the data using the moment covariance.  
Let $B$ satisfy $B B^\top = (\hat{\Sigma} + \varepsilon I_d)^{-1}$ with $\varepsilon = 10^{-12}$, and define
$\check{X}_i = B(X_i - \bar{X})$ and $\check{x} = B(x - \bar{X})$.  
The spatial depth is then
$$
D_{\mathrm{spat}}(x \mid X_{1:n})
= 1 - \Big\| \frac{1}{n}\sum_{i=1}^n s(\check{x} - \check{X}_i) \Big\|_2.
$$

\paragraph{Potential depth.} \citep{pokotylo2019classification}  
We follow the kernel-based formulation of \citet{pokotylo2019classification}, but omit whitening for improved numerical stability in high dimensions.  
Given a query point $x$ and reference sample $\{X_i\}_{i=1}^n$, define distances $r_i = \|x - X_i\|_2$.  
The bandwidth parameter $\beta > 0$ is chosen by Scott’s rule
$\beta = n^{-2/(d+4)}$, and is clamped below by a small $\varepsilon = 10^{-12}$ for stability.  
The potential depth is the average of a radial kernel:
$$
D_{\mathrm{pot}}(x \mid X_{1:n})
= \frac{1}{n}\sum_{i=1}^n K_\beta(r_i),
\qquad
K_\beta(r) = \exp(-\beta r^2)  \text{ (Gaussian kernel)}.
$$

\paragraph{Projection depth.} \citep{dyckerhoff2021approximate}  
We approximate the Stahel–Donoho outlyingness by sampling $K$ random unit directions
$u_k \in \mathbb{S}^{d-1}$ and computing
$$
D_{\mathrm{proj}}(x \mid X_{1:n})
= \frac{1}{1 + \max_{k \le K}
\frac{|\langle x, u_k \rangle - \mathrm{med}(\langle X, u_k \rangle)|}
{\mathrm{MAD}(\langle X_{1:n}, u_k \rangle)}}.
$$
Here $\langle X_{1:n}, u_k \rangle$ denotes the vector
$(\langle X_1,u_k\rangle,\ldots,\langle X_n,u_k\rangle)^\top$, and
$\mathrm{med}$ and $\mathrm{MAD}$ are the univariate median and median absolute deviation.

\paragraph{Tukey depth.} \citep{tukey1975mathematics} 
We approximate the Tukey depth using its projection-min characterization.  
Sampling $K$ random directions $u_k \in \mathbb{S}^{d-1}$, we project $X$ and $x$ onto each $u_k$, compute the univariate Tukey depth in the projected space, and take the minimum:
$$
D_{\mathrm{half}}(x \mid X_{1:n})
= \min_{k \le K} D_{\mathrm{tukey}}\big(\langle x, u_k \rangle \mid \langle X_{1:n}, u_k \rangle\big).
$$

\paragraph{Direction sampling for projection-based methods.}
Both projection depth and halfspace depth rely on random directions to approximate directional extrema.  
To balance runtime and accuracy, we use
$$
K = \operatorname{clip}\!\big(n \times 2^{\,d-1},\,100,\,50{,}000\big),
\qquad
\operatorname{clip}(x,a,b)=\min\{\max\{x,a\},b\}.
$$
This rule scales the direction budget with dimensionality while keeping computation tractable:  
at least 100 directions are used in low dimensions, and a maximum of $50{,}000$ directions caps the cost in higher dimensions.  
This setting achieves a good trade-off between approximation accuracy and computational efficiency across all datasets.

\section{Outlier detection experiment settings}
\label{app:outlier}

\paragraph{Cross-validation and model selection.}
We use a simple 5-fold scheme without nested inner loops. For outer fold
$k\in\{0,1,2,3,4\}$, fold $k$ serves as the test set, fold $(k{+}1)\bmod 5$ as
the validation set, and the remaining three folds as the training set.
For each method and fold, hyperparameters are selected on the validation set by
maximizing AUC-PRC (ties broken by AUC-ROC), after which the model is refit on
the corresponding training folds and evaluated on the held-out test fold. Final
numbers are the mean over the five test folds.

\paragraph{Metrics.}
We evaluate performance using area under the precision–recall curve (AUC-PRC)
and area under the receiver-operating characteristic curve (AUC-ROC). Because
anomalies are rare, AUC-PRC is our primary metric, and AUC-ROC is reported for
completeness. For models whose raw outputs represent inlier scores (e.g., KDE
log-density, IF/LOF decision scores), we negate the scores so that larger values
consistently indicate stronger outlierness before computing the AUC metrics.

\paragraph{Hyperparameter grids.}
We use standard \texttt{scikit-learn} implementations with the hyperparameter
grids listed below. Unless otherwise specified, all remaining parameters are
kept at their \texttt{scikit-learn} default values.

\begin{itemize}[leftmargin=*]
  \item \textbf{IF} (\texttt{sklearn.ensemble.IsolationForest}):
  $n_{\mathrm{estimators}} \in \{100,200,$ $400\}$;
  $max\_features \in \{1.0,0.75,0.5\}$;
  $max\_samples = \texttt{"auto"}$;
  $bootstrap = \texttt{False}$.

  \item \textbf{LOF }(\texttt{sklearn.neighbors.LocalOutlierFactor}):
  $n_{\mathrm{neighbors}} \in \{10,20,35,50\}$;
  Minkowski order $p \in \{1,2\}$ ($L_1$ or $L_2$);
  $leaf\_size = 30$ (fixed);
  $novelty = \texttt{True}$.

  \item \textbf{OCSVM} (\texttt{sklearn.svm.OneClassSVM}):
  $\gamma \in \{\texttt{"scale"},\texttt{"auto"},10^{-3},$ $10^{-2},10^{-1}\}$;
  $\nu \in \{0.01,$ $0.05,0.1,0.2\}$;
  $max\_iter = 10^5$.

  \item \textbf{KDE}   (\texttt{sklearn.neighbors.KernelDensity}):
  $kernel \in \{\texttt{"gaussian"},\texttt{"epanechnikov"}\}$;
  $bandwidth \in \{0.1,0.2,0.5,1.0,2.0\}$;
  $metric = \texttt{"euclidean"}$.

  \item \textbf{FUSE}: We use the architecture from
Supplement~\ref{app:net-arch} with shared-encoder width $(128,128)$. Distance metric $\in \{1,2\}$ ($L_1$ or $L_2$);
  DSM noise scale $\eta \in \{0.1, 0.2, 0.5,1.0,2.0\}$;
  interpolation parameter $t \in \{0.1,0.2,\dots,0.9\}$;
  number of anchors $=64$;
  DSM resamples per epoch $=8$;
  batch size $=128$;
  epochs $=30$;
  learning rate $=10^{-3}$. The endpoints $t{=}0$ and $t{=}1$ correspond to
the Global and Local variants, respectively, and are reported separately. 
\end{itemize}


\end{document}

%% file: math_commands.tex
\usepackage{amsmath,amsfonts,bm}

\def\eqref#1{equation~\ref{#1}}

\def\1{\bm{1}}

\DeclareMathAlphabet{\mathsfit}{\encodingdefault}{\sfdefault}{m}{sl}
\SetMathAlphabet{\mathsfit}{bold}{\encodingdefault}{\sfdefault}{bx}{n}

